\pdfoutput=1

\documentclass[11pt]{article}

\usepackage{EACL2023}

\usepackage{times}
\usepackage{mathptmx}
\usepackage{caption}
\usepackage{subcaption}
\usepackage{latexsym}
\usepackage{color}

\usepackage[T1]{fontenc}

\usepackage[utf8]{inputenc}

\usepackage{microtype}

\usepackage{inconsolata}

\usepackage[utf8]{inputenc}
\usepackage[T1]{fontenc}
\usepackage{hyphenat}
\usepackage{xspace}
\usepackage{amsmath}
\usepackage{amsfonts}
\usepackage{hyperref}
\usepackage{url}
\usepackage{booktabs}
\usepackage{multirow}
\usepackage{makecell}
\usepackage{caption}
\usepackage{minibox}
\usepackage{bbm}
\usepackage{graphicx}
\usepackage{balance}
\usepackage{mathtools}
\usepackage{color}
\usepackage{marvosym}
\usepackage{ifthen}
\usepackage{textcomp}
\usepackage{enumitem}
\usepackage{verbatim}
\usepackage{algorithm}
\usepackage{numprint}
\usepackage{balance}

\usepackage{amsthm}
\theoremstyle{plain}

\newcommand{\chatoDisplayMode}[1]{#1}

\definecolor{MyRed}{rgb}{0.6,0.0,0.0} 
\definecolor{MyBlack}{rgb}{0.1,0.1,0.1} 
\newcommand{\inred}[1]{{\color{MyRed}\sf\textbf{\textsc{#1}}}}
\newcommand{\frameit}[2]{
  \begin{center}
  {\color{MyRed}
  \framebox[.9\columnwidth][l]{
    \begin{minipage}{.85\columnwidth}
    \inred{#1}: {\sf\color{MyBlack}#2}
    \end{minipage}
  }\\
  }
  \end{center}
}

\newcommand{\note}[2][]{\chatoDisplayMode{\def\@tmpsig{#1}\frameit{{\Pointinghand} Note}{#2\ifx \@tmpsig \@empty \else \mbox{ --\em #1}\fi}}}
\newcommand{\todo}[2][]{\chatoDisplayMode{\def\@tmpsig{#1}\frameit{{\Writinghand} To-do}{#2\ifx \@tmpsig \@empty \else \mbox{ --\em #1}\fi}}}

\newcommand{\abbrevStyle}[1]{#1}

\newcommand{\ie}{\abbrevStyle{i.e.}\xspace}
\newcommand{\eg}{\abbrevStyle{e.g.}\xspace}

\newcommand{\Secref}[1]{Sec.~\ref{#1}}

\newcommand{\Tabref}[1]{Table~\ref{#1}}
\newcommand{\Figref}[1]{Fig.~\ref{#1}}

\newcommand{\Appref}[1]{Appendix~\ref{#1}}

\newcommand{\xhdr}[1]{\vspace{1.7mm}\noindent{{\bf #1.}}}

\newcommand{\textcite}[1]{\citeauthor{#1} \shortcite{#1}}

\newcommand{\hide}[1]{}

\makeatletter
\newcommand{\iffont}[2]{\ifthenelse{\equal{\f@family}{#1}}{#2}{}}
\makeatother

\title{Language Model Decoding as Likelihood--Utility Alignment}

\DeclareSymbolFont{extraup}{U}{zavm}{m}{n}
\DeclareMathSymbol{\microsoft}{\mathalpha}{extraup}{81}
\DeclareMathSymbol{\epfl}{\mathalpha}{extraup}{83}
\DeclareMathSymbol{\psl}{\mathalpha}{extraup}{84}

\author{
Martin Josifoski,$^{\epfl}$
Maxime Peyrard,$^{\epfl}$ 
Frano Rajic,$^{\epfl}$
Jiheng Wei,$^{\psl}$ 
\AND
Debjit Paul,$^{\epfl}$ 
Valentin Hartmann,$^{\epfl}$
Barun Patra,$^{\microsoft}$
Vishrav Chaudhary,$^{\microsoft}$
\AND \vspace{5pt}
Emre K\i{}c\i{}man,$^{\microsoft}$
Boi Faltings,$^{\epfl}$
Robert West$^{\epfl}$ \\
    $^{\epfl}$EPFL \quad $^{\microsoft}$Microsoft Corporation \quad $^{\psl}$PSL University \\
    {\{martin.josifoski, maxime.peyrard, robert.west\}@epfl.ch} \\
  }

\begin{document}
\maketitle
\begin{abstract}

A critical component of a successful language generation pipeline is the \emph{decoding algorithm}.
However, the general principles that should guide the choice of a decoding algorithm remain unclear. Previous works only compare decoding algorithms in narrow scenarios, and their findings do not generalize across tasks. 
We argue that the misalignment between the model's \emph{likelihood} and the task-specific notion of \emph{utility} is the key factor to understanding the effectiveness of decoding algorithms.
To structure the discussion, we introduce a taxonomy of misalignment mitigation strategies (MMSs), providing a unifying view of decoding as a tool for alignment. The MMS taxonomy groups decoding algorithms based on their implicit assumptions about likelihood--utility misalignment, yielding general statements about their applicability across tasks. 
Specifically, by analyzing the correlation between the likelihood and the utility of predictions across a diverse set of tasks, we provide empirical evidence supporting the proposed taxonomy and a set of principles to structure reasoning when choosing a decoding algorithm.
Crucially, our analysis is the first to relate likelihood-based decoding algorithms with algorithms that rely on external information, such as value-guided methods and prompting, and covers the most diverse set of tasks to date. Code, data, and models are available at \href{https://github.com/epfl-dlab/understanding-decoding}{https://github.com/epfl-dlab/understanding-decoding}.

\end{abstract}

\section{Introduction}
\label{sec:introduction}

\begin{figure}[t]
     \centering
         \includegraphics[width=0.70\columnwidth]{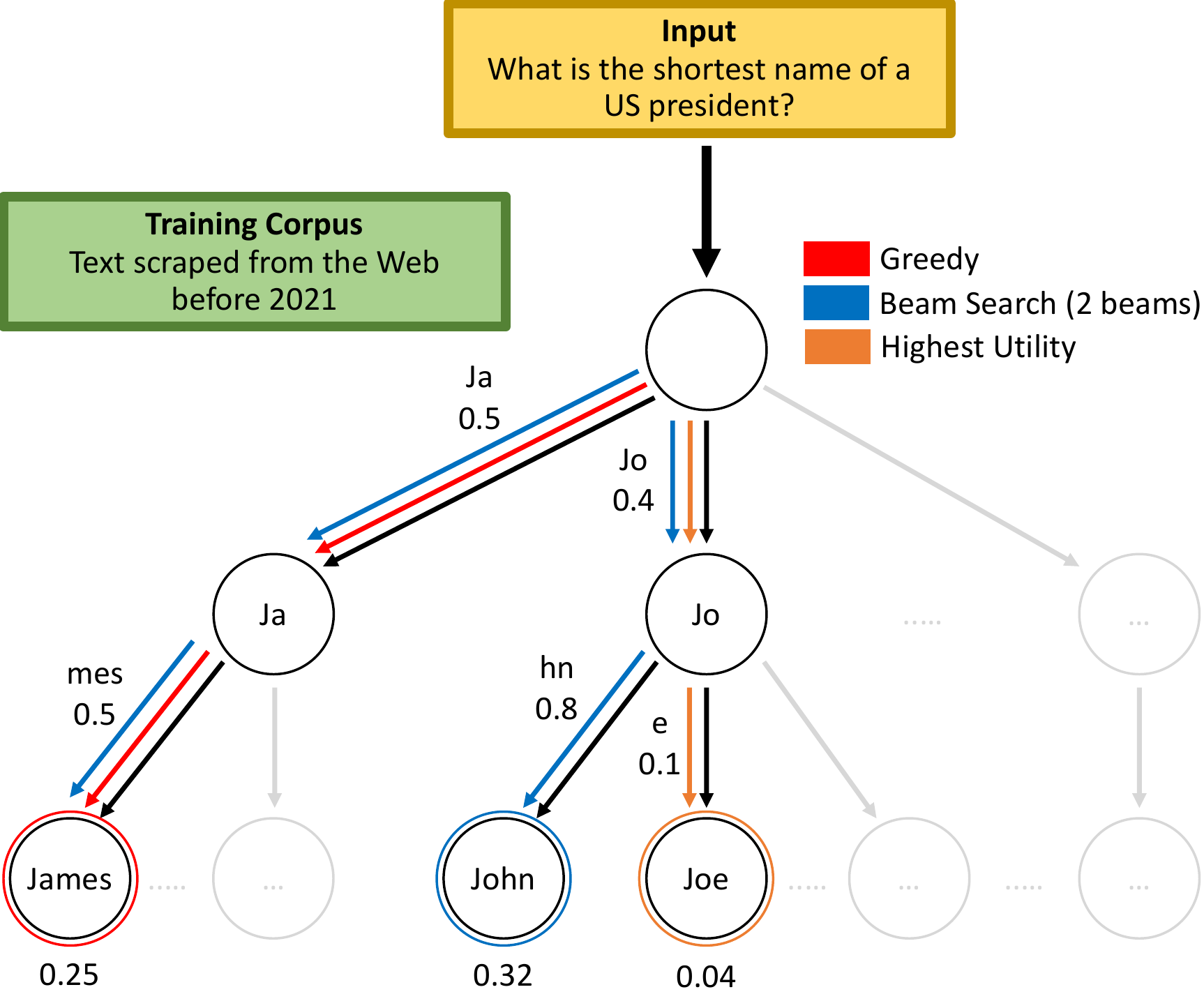}
         \caption{\textbf{Example of likelihood--utility misalignment.} Imagine a fictional LM trained before Joe Biden became the US president. The input asks for the shortest name of a US president. After Joe Biden's inauguration, this is `Joe', but before, it was `John'. Greedy search returns `James' since its first token `Ja' has the highest likelihood. Beam search manages to find the highest likelihood sequence `John'. Both fail to find the correct answer `Joe' with the highest utility since `Joe' has a very low likelihood.}
         \label{fig:diagram}
         \vspace{-1em}
\end{figure}

Large transformer-based \textit{language models} (LMs) have been pushing the boundaries on tasks ranging from natural language generation \cite{gpt2} to information extraction \cite{josifoski-etal-2022-genie}, theorem proving \cite{DBLP:journals/corr/abs-2009-03393}, code generation \citep{lm4code}, and even protein generation \citep{Ferruz2022ProtGPT2IA}. At inference time, these models rely on a decoding algorithm to generate an output. The goal of decoding algorithms is to select an output of high utility from the exponentially large output space. In contrast to the generic language modeling training objective, which is based on the data likelihood, the notion of utility is task-specific. The potential gap between the two can create a \emph{misalignment} between  \emph{model likelihood} and  \emph{task utility}; see \Figref{fig:diagram} for an illustration of this concept.

Indeed, across different tasks, researchers noticed that high likelihood is often not associated with desired properties of the output \cite{stahlberg-byrne-2019-nmt, zhang-etal-2021-trading, klein-etal-2017-opennmt}. Naturally, this has led to the development of decoding strategies aimed at mitigating this problem. In the context of natural language generation (NLG), Nucleus Sampling (\ie, top-$p$) \cite{DBLP:conf/iclr/HoltzmanBDFC20} has been proposed to avoid dull or degenerate text. Similarly, in the context of machine translation (MT), solutions ranging from simple ad-hoc tweaks like enforcing a minimal sequence length \cite{stahlberg-byrne-2019-nmt} to leveraging a value model to directly optimize for utility in decoding \cite{leblond-etal-2021-machine} have been developed. These methods for alignment are task-specific and have been tested only in narrow domains, making it difficult for practitioners to compare them.

Recently, \newcite{meister2022on} and \newcite{https://doi.org/10.48550/arxiv.2203.15721} explored the 
likelihood--utility misalignment across tasks.  
However, these studies still 
largely focus on a group of similar tasks --- NLG tasks --- and, more crucially,
do not include decoding strategies that make use of external sources of information at inference time.
Therefore, a general framework to structure our thinking about decoding algorithms is still missing.

Our work makes the first step towards filling this gap. 
We propose a unified perspective of decoding as a tool for mitigating the likelihood--utility misalignment without modifications to the model. Looking at decoding through this functional lens, in \Secref{sec:taxonomy}, we provide a taxonomy of misalignment mitigation strategies (MMSs). The taxonomy groups decoding algorithms based on the implicit assumptions about likelihood--utility misalignment that need to hold for them to be effective.

Equipped with this taxonomy, we conduct a comprehensive empirical analysis in which we choose a representative set of decoding algorithms and a representative set of tasks to cover the relevant types of misalignment. We identify three main sources of misalignment: \textit{training imperfections} (finite dataset, differentiable surrogate loss), \textit{distribution shift}, and changes in the model's intended usage, which we call \emph{utility drift}.
Then, we measure the likelihood--utility misalignment across tasks (RQ1) by estimating: the correlation between likelihood and utility for the generated outputs after decoding (RQ1-a) and the correlation between likelihood and utility among candidate outputs explored by the decoding algorithm (RQ1-b). We proceed by investigating the benefits of decoding algorithms that leverage external information at inference (RQ2). Finally, we experiment with large generalist LMs (LLMs) and show that prompting can be seen as means for improving the alignment at inference time (RQ3). 

Our experiments reveal that: (i) When no distribution shift or utility drift happens, decoding based solely on the likelihood is enough to provide high utility, i.e., likelihood is a strong predictor of utility. (ii) In such cases, there is no significant difference between different kinds of decoding algorithms, and we would recommend keeping beam search. (iii) In the presence of distribution shift or utility drift, value-guided beam search is both an effective and efficient decoding algorithm that leverages a value model at inference time to fix misalignment. Finally, (iv) for LLMs, prompting is a mechanism that sets the model in a state where the likelihood is well-aligned with the utility. This perspective provides a tentative explanation for the empirical success of prompting LLMs.

This work studies the fundamental problem in decoding, which involves a complex interaction between models, tasks, and data. Our unifying conceptual framework (the MMS taxonomy), accommodating all known decoding algorithms, enables the systematic study of decoding in a considerably broader scope than previously. By sharing the taxonomy and open-sourcing our implementations, we hope to pave the way for a more structured discussion in the future.

\section{Background}
\label{sec:background}
\xhdr{High Utility Is the Goal}
For a task $t$ and input $x$, the utility function assigns a score $u_t (y | x)$ to each element $y$ in the output space $\mathcal{Y}$.
This score quantifies the goodness, or quality, of the output $y$ with respect to a specific input $x$. For instance, in translation, the utility quantifies the extent to which the output conveys the same message as the input. For question answering, the utility simply quantifies the correctness of the answer. These task-specific notions of utility are operationalized in the evaluation metrics. The development of evaluation metrics that correlate with the human-defined notion of utility is a very active research area \cite{10.1145/3485766} and beyond the scope of this work. In our analysis, we use the canonical evaluation metric of each task as the utility function. For partially-decoded sequences, the utility can be approximated using a \emph{value function} (see \Secref{ssec:vgbs}).

Given some input $x$, an \emph{ideal model} would generate the element from the output space corresponding to the highest utility score:
$\operatorname*{argmax}_{y \in \mathcal{Y}} u_t(y\;|\;x)$.

Unfortunately, most of the practically relevant utility functions are not amenable to optimization, forcing us to work with proxy functions, such as the canonical likelihood.

\xhdr{Language Models}
A language model corresponds to a probability distribution $p$ over $y$ $\in \mathcal{Y}$, where $\mathcal{Y}$ is the set of all sequences that can be constructed using a vocabulary $\mathcal{V}$. 
In this work, we focus on conditional LMs $p(\cdot | x)$. Usually, these conditional distributions are modeled autoregressively (parametrized by $\theta$): 
    $p_\theta(y\mid x) = \prod_{i=1}^{|y|} p_\theta(y_i\mid y_{<i}, x).$
The model is trained to maximize the target sequence's conditional log-likelihood with teacher forcing, using the cross-entropy loss
$\mathcal{L}(\theta)=-\sum_{(x, y) \in \mathcal{D}} \log p_\theta(y\;|\;x),$
where $\mathcal{D}$ is the training corpus~\citep{sutskever2011generating, sutskever2014sequence}.

Once an LM is trained, it provides a next-token probability distribution across the output vocabulary. Decoding algorithms define how tokens are chosen during generation.

\section{Proposed MMS Taxonomy}
\label{sec:taxonomy}
In this section, we propose a taxonomy of misalignment mitigation strategies (MMSs). This work focuses on decoding-based MMSs that mitigate the misalignment without modifying the model. As a primary signal, they rely on the model's likelihood. However, additional components (\eg, value model, knowledge base, etc.) can be leveraged. Decoding algorithms, which we define as procedures that take in an input --- and potentially some context (\eg, a prompt with a task description or examples) --- and return a sequence from the output space, can be seen as specific implementations of an MMS.
Apart from fixing the misalignment problem at inference time, it is also possible to retrain or finetune the model with newly collected data that better reflect the intended utility and target testing distribution. We leave the detailed treatment of this part of the taxonomy for future work. See the Limitations section at the end of the writing for further discussion of these alternatives.

\subsection{Greedy Likelihood-Based Strategy}
Given the LM's probabilistic formulation, one could strive to select the most likely sequence under the model: $\operatorname*{argmax}_{y \in \mathcal{Y}} p_\theta(y\;|\;x).$
However, due to the exponentially large state space, this optimization problem is intractable.

The class of algorithms following the greedy likelihood-based MMS approximate the intractable argmax
by following the greedy heuristic of making {\em locally} optimal choices at each decoding step w.r.t.\ the {\em likelihood} under the language model. However, reaching a globally optimal solution may require locally sub-optimal steps. When this happens, we say that the likelihood landscape is {\em greedy adversarial} \cite{meister-etal-2020-beam}.
For a likelihood model that is not greedy adversarial, greedy heuristics will retrieve the highest-likelihood solution. Therefore, the algorithms' effectiveness depends on the likelihood--utility alignment.

Contrarily, greedy decoding algorithms may fall arbitrarily short of the global maximum for likelihood models that are greedy adversarial. Indeed, greedy decoding algorithms implicitly optimize a different objective function --- a {\em tampered} version of the {\em likelihood} objective in which a term that encourages locally optimal solutions is added \cite{meister-etal-2020-beam}. Therefore, the ability of greedy decoding algorithms to retrieve high-utility sequences even from a likelihood model that is perfectly aligned with the utility 
is inversely proportional to how greedy adversarial the likelihood landscape is. In some cases, the particular bias induced by the greedy heuristic {\em mitigates} the likelihood--utility {\em misalignment}, and makes the tampered likelihood objective better aligned with the utility than the original \cite{meister-etal-2020-beam, 10.1162/tacl_a_00536, su2022a}. 

The decoding algorithms in this category can be further divided into two subgroups: (i) deterministic ones, such as greedy search (GS) and beam search (BS); and (ii) stochastic ones, such as top-$k$ sampling \citep{fan-etal-2018-hierarchical}, top-$p$ sampling  \citep{DBLP:conf/iclr/HoltzmanBDFC20}, and stochastic beams (SB) \cite{pmlr-v97-kool19a}. For more details, see Appendix \ref{appendix:taxonomy_greedy}.

\subsection{Greedy Likelihood-Based Strategy with Pruning}

An understated fact in the literature is that even for tasks for which the canonical decoding algorithms (\eg, BS) perform well, a non-negligible portion of the performance relies on some bespoke, ad-hoc tweaks on the likelihood scores \cite{stahlberg-byrne-2019-nmt}. 
These tweaks are usually based on either: 
(i) post-hoc observations that likelihood-based decoded sequences often contain specific undesirable patterns (\eg, empty or short sequences, repetitive patterns, etc.); or
(ii) problem-specific knowledge about the utility landscape suggests that high-utility sequences have a specific property, which can be explicitly enforced by the decoding strategy (\eg, sequences should correspond to triplets of elements from a predefined set). 
Conceptually, all these tweaks employ mechanisms that {\em discourage} the generation of {\em high-likelihood} patterns that are known (or expected) to be associated with {\em low utility}.%

This category includes: (i) decoding algorithms with ad-hoc heuristics such as the $n$-gram repetition penalty \citep{klein-etal-2017-opennmt}; (ii) constrained beam search (CBS) \cite{scholak-etal-2021-picard, de-cao-etal-2022-multilingual, josifoski-etal-2022-genie}; (iii) NeuroLogic \cite{lu-etal-2022-neurologic}. For more details, see Appendix \ref{appendix:taxonomy_greedy_with_pruning}.

\subsection{Greedy Likelihood- and Value-Based Strategy}
\label{ssec:vgbs}
Heuristics such as the ones used in the previous category can capture some properties associated with high utility but are limited to properties that can be easily expressed explicitly. While the utility function is generally defined for complete sequences only, to guide decoding algorithms, one can rely on the general concept of a value model. A value model approximates, for partially decoded sequences, the expected utility of the final sequence if the decoding keeps following the same policy.
The most prominent algorithm in this category is value-guided beam search (VGBS) \cite{NIPS2017_2b24d495, DBLP:conf/cvpr/RenWZLL17, krishna-etal-2022-rankgen}.
It uses a greedy strategy similar to BS but selects the next token using a linear combination of the LM’s likelihood and the value model's scores. See Appendix \ref{appendix:taxonomy_value_based} for details.

\subsection{Simulation-Based Strategy}

Even though VGBS considers both likelihood and value, it remains greedy by only looking one step ahead. Simulation-based decoding algorithms explore further into the future before making the next decision. When the value landscape is complex and constructing a good value model is hard, such strategies with a look-ahead may become particularly effective. By turning the knob controlling the number of simulations, one can trade off computational efficiency for obtaining better value estimates. Monte\hyp Carlo Tree Search (MCTS) is the canonical example of simulation-based tree exploration informed by a value model. For details, see Appendix \ref{appendix:taxonomy_simulation_based}.

\subsection{Prompting-Based Strategy}
The decoding algorithms described in the last two sections address the likelihood--utility misalignment post-hoc. An alternative is to change the conditioning of the model's probability distribution such that the misalignment never happens. The effort now goes into choosing a context that aligns the likelihood landscape with the task-specific utility. The strength of this class of decoding algorithms lies in the fact that they can readily be applied to a new task without requiring any modifications to the model or increasing the computation cost of inference (beyond the processing of the prompts' tokens).
However, they only work for 
large generalist LMs  \cite{DBLP:journals/corr/abs-2204-02311}. 
The most prominent members are the few-shot (FS) and the chain-of-thought (CoT) prompting methods, described in Appendix \ref{appendix:taxonomy_prompting}.

\section{Experimental Setup}
\label{sec:experimental_setup}
\subsection{Research Questions}

In contrast to previous works that have studied the misalignment problem in constrained settings, we propose quantifying it in a unified and large-scale analysis across tasks (RQ1). We investigate the benefits of a diverse set of previously proposed solutions to the misalignment problem. Our study includes value-guided approaches (RQ2) and prompting (RQ3), covering each class of the MMS taxonomy with at least one representative. Specifically, 
we ask the following research questions.

\textbf{RQ1: How correlated are the utility and the likelihood across tasks?}
As argued in \Secref{sec:taxonomy}, greedy likelihood-based strategies only require the likelihood to be a strong predictor of utility. We investigate whether this holds across tasks. Specifically, we measure two important aspects of the likelihood--utility alignment:
(a) \textbf{Post-decoding alignment}. For each data point, the decoding algorithm chooses one output; we measure the likelihood and utility of the prediction and analyze their relation. Is high likelihood associated with high utility in the same way across tasks?
(b) \textbf{During-decoding alignment}. Decoding algorithms typically explore a set of high-scoring candidates (e.g., BS returns one candidate per beam). We measure the correlation between likelihood and utility among these candidate outputs to analyze the likelihood landscape of the model.

\begin{table*}[!ht]
\centering
\small 
\scalebox{0.78}{{
\begin{tabular}{@{}l|l|l|l|l@{}}
\toprule
{\bf Tasks} & {\bf Utilities}& {\bf Misalignment Types} & {\bf Model} & {\bf Dataset} \\
\midrule
Closed Information Extraction (cIE)& {F1 score} [M] & {TI} & \href{https://github.com/epfl-dlab/genie}{GenIE (BART)} & \href{https://huggingface.co/datasets/Babelscape/rebel-dataset}{REBEL}\\
Machine Translation (MT) & BLEU [M] & {TI, DS} & 
\href{https://huggingface.co/facebook/mbart-large-50-one-to-many-mmt}{mBART50} &
{\href{https://huggingface.co/datasets/wmt14}{WMT14}}\\
Non-Toxic Text Generation (NTTG)  & {Non-Toxicity} [T] & {TI, DS, UD} & \href{https://huggingface.co/gpt2}{GPT2} & {\href{https://allenai.org/data/real-toxicity-prompts}{RTP}} \\
Non-Soluble Protein Generation (NSPG) & {Non-Solubility} [T]& {TI, DS, UD} & \href{https://huggingface.co/nferruz/ProtGPT2}{ProtoGPT2} & {\href{https://huggingface.co/datasets/lightonai/SwissProt-EC-leaf}{SwissProt}} \\
Sports Understanding & Solve Rate [M] & TI, DS, UD  & \href{https://www.microsoft.com/en-us/research/blog/turing-nlg-a-17-billion-parameter-language-model-by-microsoft/}{MT-NLG 530B} & \href{https://github.com/google/BIG-bench/tree/main/bigbench/benchmark_tasks/sports_understanding}{Sports} \\
\bottomrule
\end{tabular}}}
\caption{\textbf{Overview of the tasks.}
Utility functions are categorized into: (a) [M]: Metric-based and (b) [T]: Trained model-based. The three misalignment types are TI: Training imperfection, DS: Distribution shift, UD: Utility drift. }\label{tab:data_description}
\end{table*}

\textbf{RQ2: How effective are value-guided MMSs?} 
In particular, we investigate the benefit of value-guided decoding algorithms as a function of the value model's quality.

\textbf{RQ3: Is prompting an MMS?} We investigate the efficacy of prompting as a likelihood--utility alignment tool for generalist LMs (LLMs).

\subsection{Tasks and Datasets} \label{sec:4.2_task_datasets}
To organize the discussion, we propose a simple classification of the sources of misalignment: (a) \textbf{Training imperfections (TI)}, when the model is trained on a different objective than the true utility, because of the finite size of the dataset and the approximation error in training (\eg, via stochastic gradient descent); (b) \textbf{Distribution shift (DS)}, when the training and testing data distributions differ; 
(c) \textbf{Utility drift (UD)}, when the utility used in development differs from the utility at test time. 

While we can expect TI to affect all machine learning tasks (due to the finiteness of datasets and approximations resulting from gradient-based training), DS and UD are task-specific. DS typically occurs when the distribution of the data changes between the training and testing scenarios. UD occurs when the notion of utility changes, \ie, the labels for the same data points are changing.

We carefully selected a variety of generation tasks covering (a) different \textit{notions of utility}, and (b) different expected types of \textit{(mis)alignment between utility and likelihood}.
Table \ref{tab:data_description} gives a high-level overview of these tasks, their utility functions, and associated datasets.\footnote{For more details about the models, data, and utility functions, see \Appref{app:data}.}
In closed information extraction, the training and testing data come from the same distribution, and we expect only TI-type of misalignment. 
In machine translation, the mBART model is pretrained on a different dataset, inducing some DS as the texts used for training may come from different domains. For non-toxic text and non-soluble protein generation, the task definition changed from generating low perplexity sequences to generating non-toxic sequences. Therefore, UD is expected to be the main driver of misalignment. Similarly, for the sports understanding task, since MT-NLG was not trained for this specific task, we also expect UD to be the main source of misalignment, but here VGBS and MCTS are too expensive due to the size of the LM. Instead, we use this setting to investigate prompting-based MMSs.

\subsection{Decoding Algorithms}
To cover the full space of MMSs, we experiment with at least one representative from each class from the taxonomy defined in Sec.~\ref{sec:taxonomy}. From the Greedy Likelihood-based category, we include the canonical GS and BS, as well as the sampling-based SB. From the Greedy Likelihood-based Strategies with pruning, we use CBS. VGBS and MCTS are representatives of the Greedy Likelihood- and Value-based, and Simulation-based decoding classes, respectively. For prompting, we consider the FS and CoT methods. The hyper-parameters for each algorithm are given in \Appref{app:hyperparam}. \Appref{app:complexity}, provides a complexity analysis in terms of LM and value model calls.

\begin{figure*}[ht!]
     \centering
     
     \rotatebox[origin=c]{90}{\footnotesize Greedy Search}
     \begin{subfigure}[c]{0.2375\textwidth}
         \centering
         \includegraphics[width=.88\textwidth]{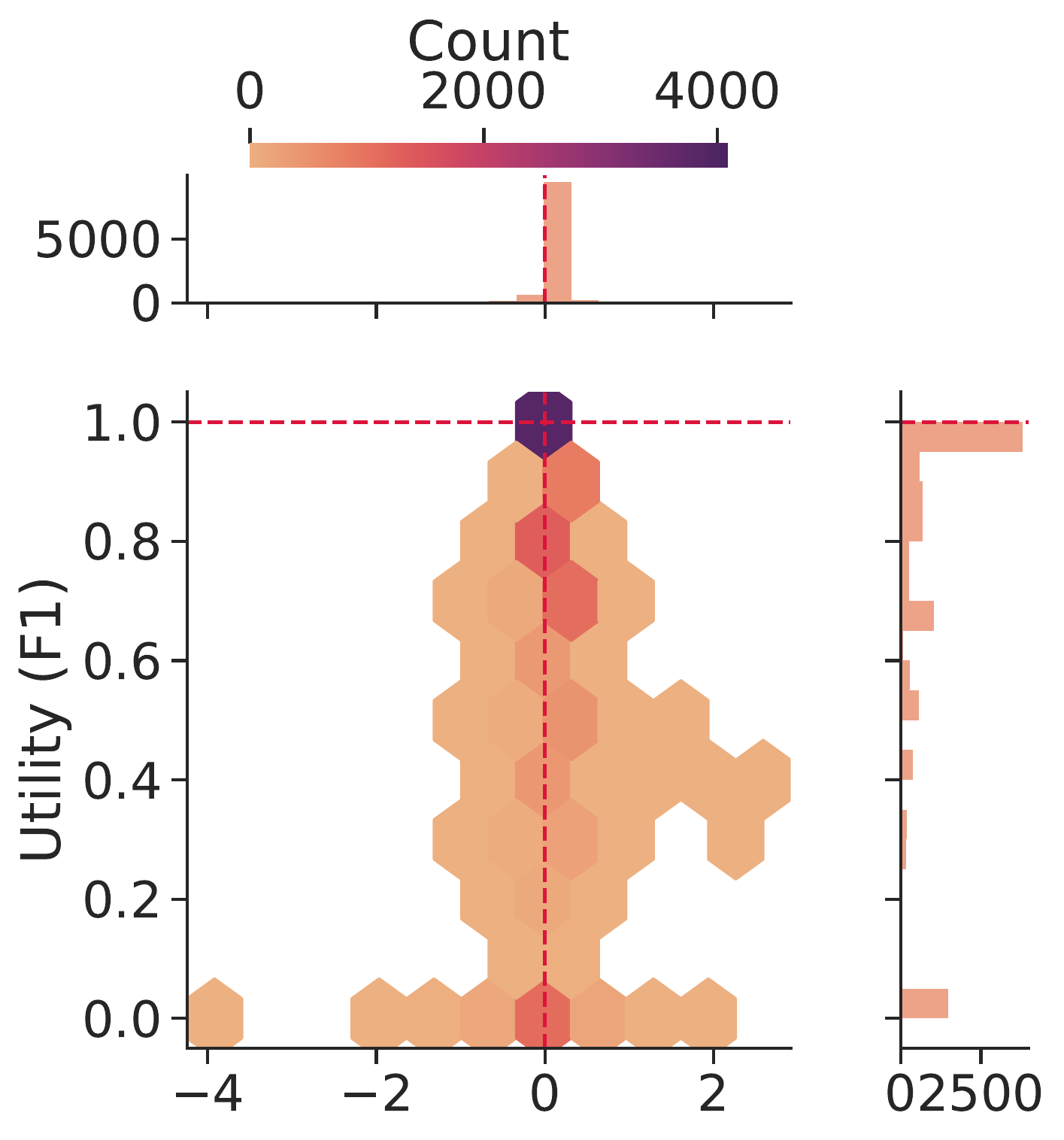}
         \label{fig:rebel_greedy}
     \end{subfigure}
     \begin{subfigure}[c]{0.2375\textwidth}
         \centering
         \includegraphics[width=.88\textwidth]{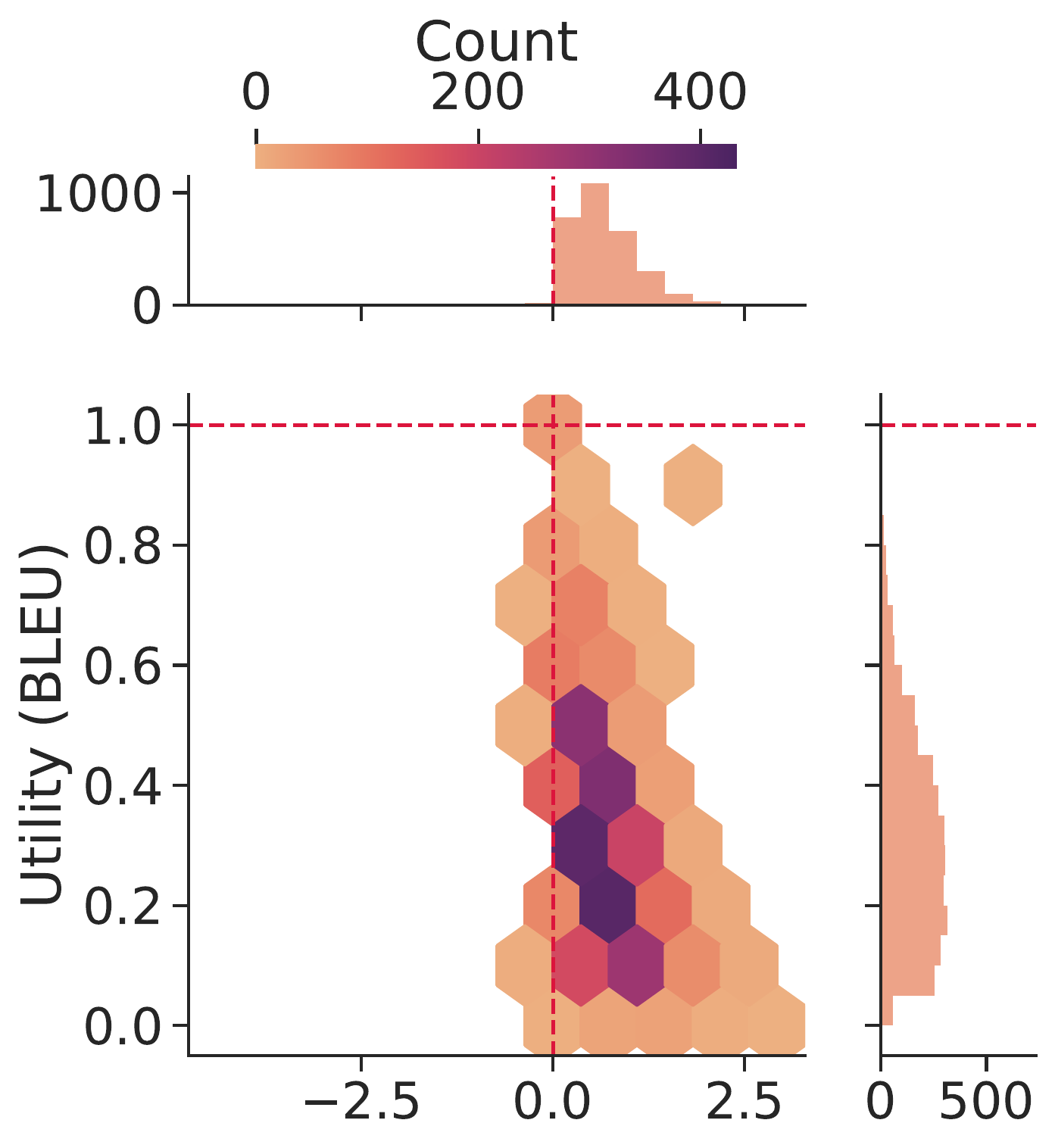}
         \label{fig:wmt14_greedy}
     \end{subfigure}
     \begin{subfigure}[c]{0.2375\textwidth}
         \centering
         \includegraphics[width=.88\textwidth]{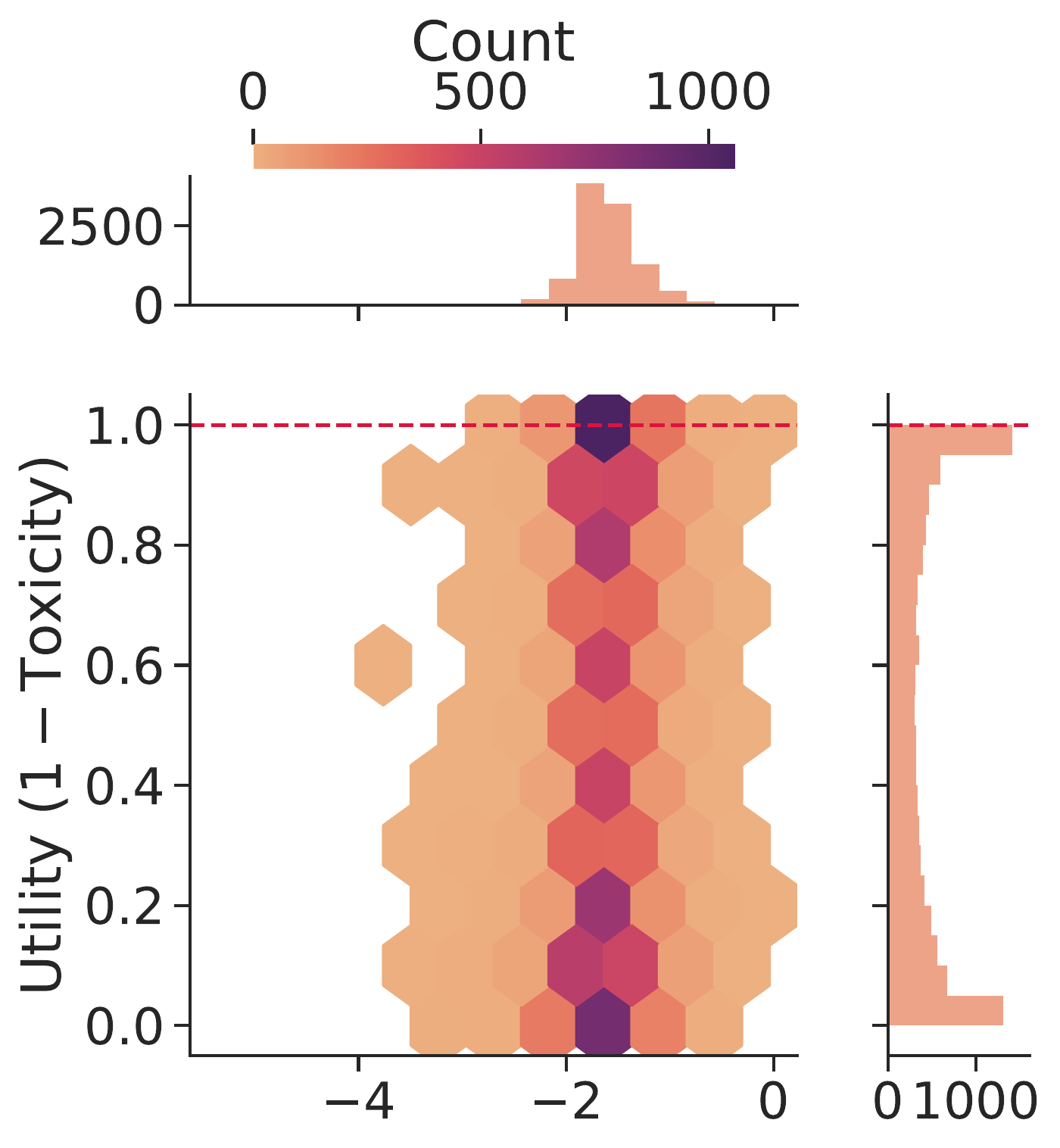}
         \label{fig:rtp_greedy}
     \end{subfigure}
     \begin{subfigure}[c]{0.2375\textwidth}
         \centering
         \includegraphics[width=.88\textwidth]{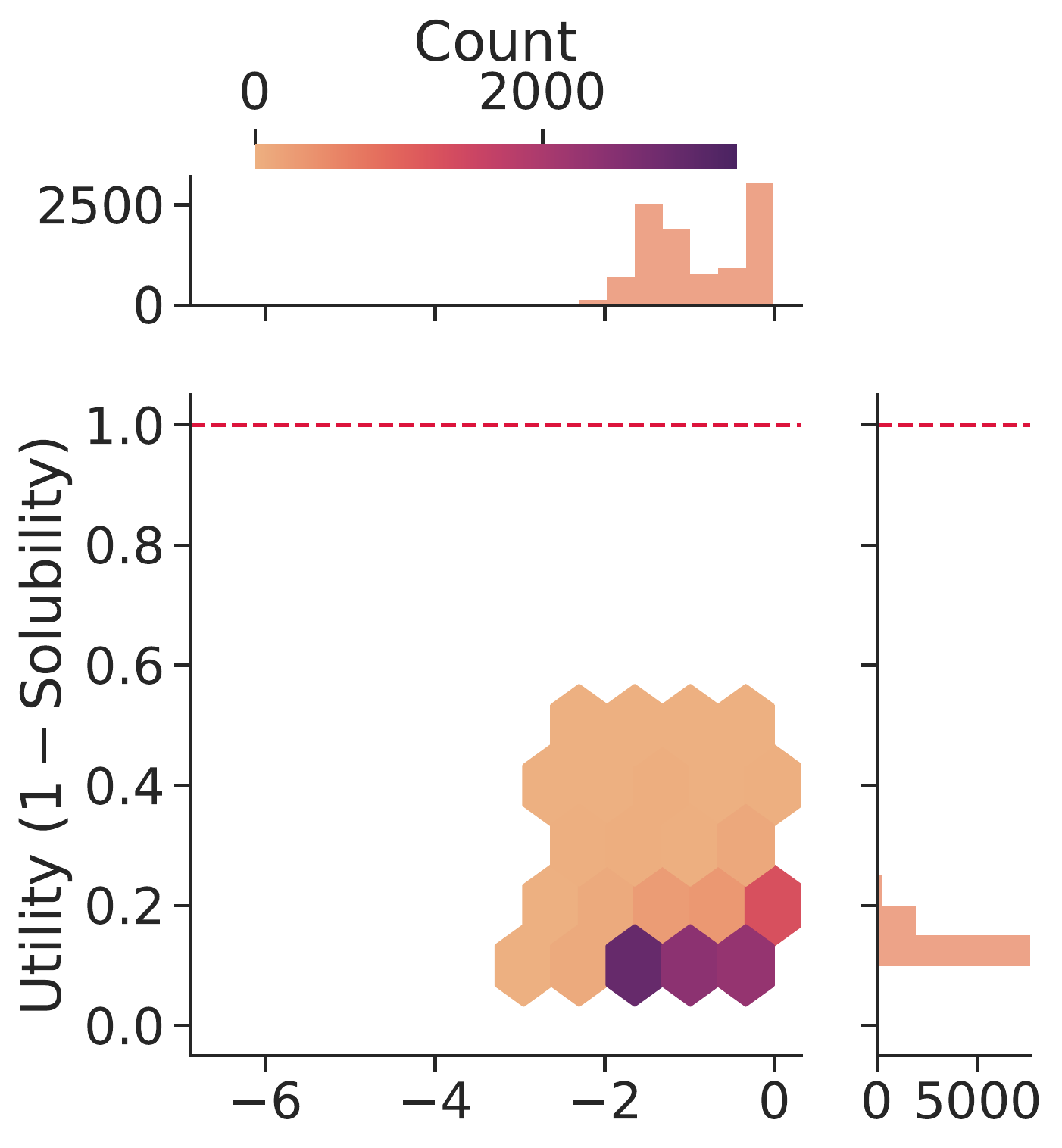}
         \label{fig:prot_greedy}
     \end{subfigure}
     
    \rotatebox[origin=c]{90}{\footnotesize Beam Search}
     \begin{subfigure}[c]{0.2375\textwidth}
         \centering
         \includegraphics[width=.88\textwidth]{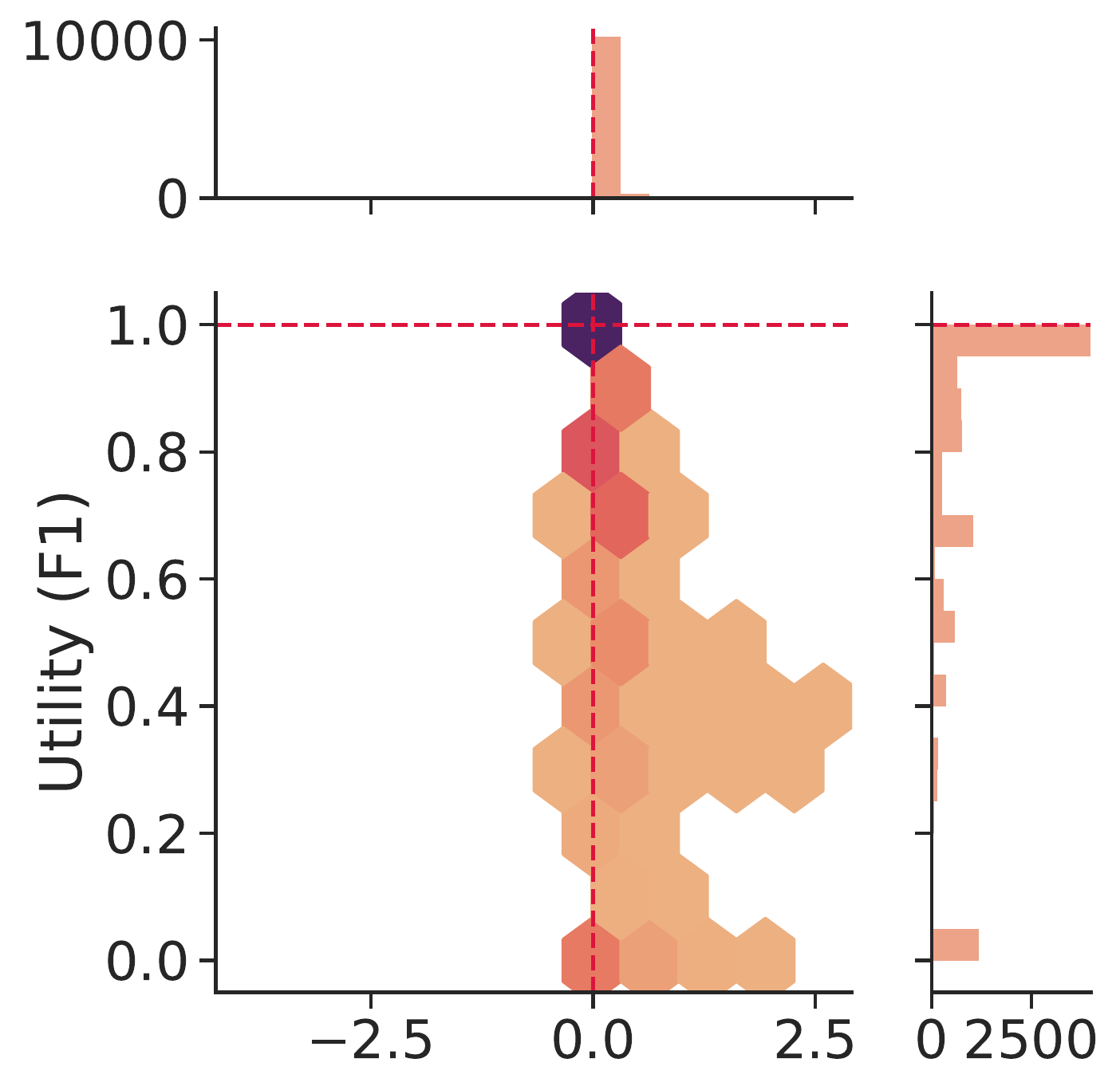}
         \label{fig:rebel_bs}
     \end{subfigure}
     \begin{subfigure}[c]{0.2375\textwidth}
         \centering
         \includegraphics[width=.88\textwidth]{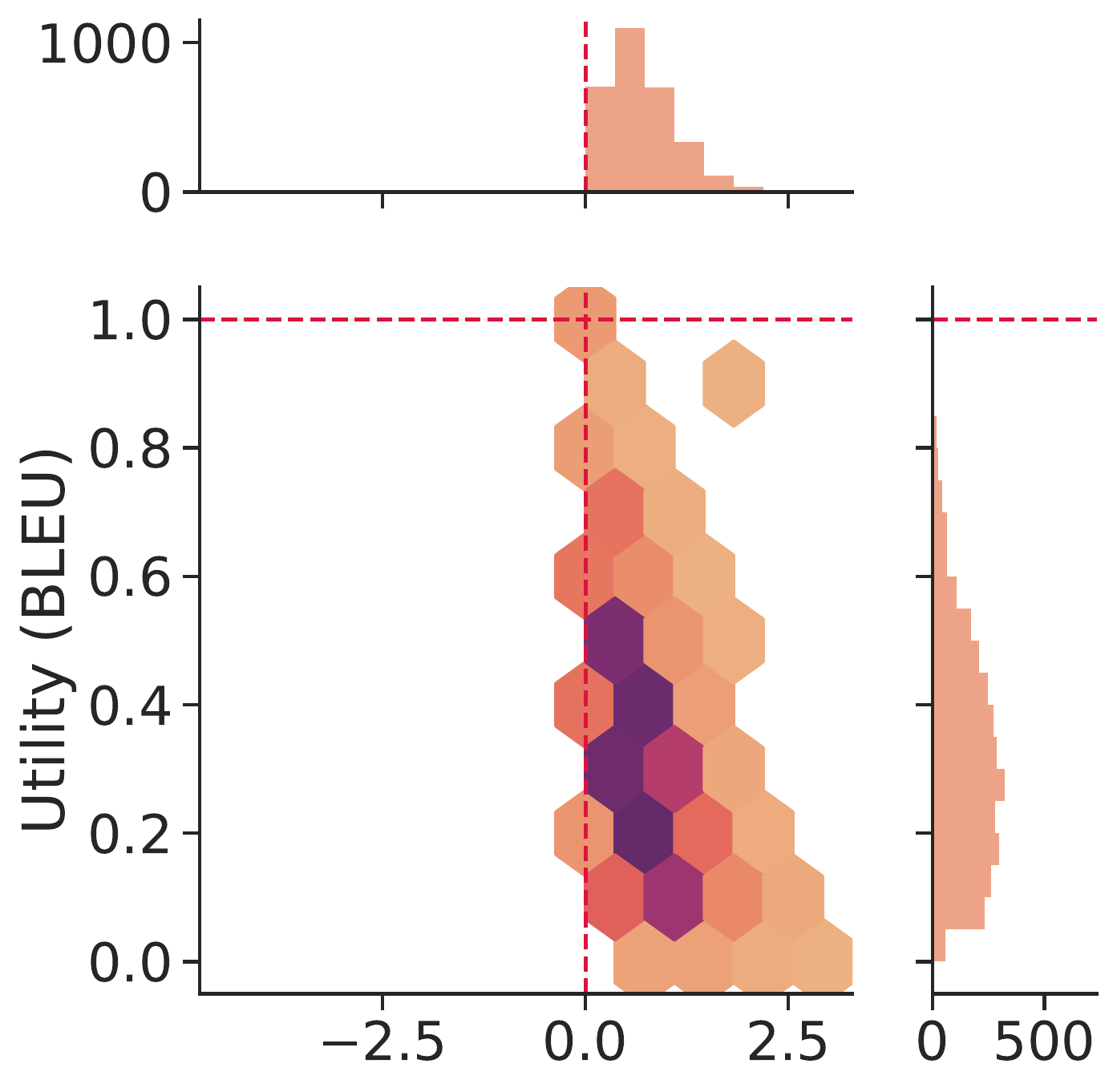}
         \label{fig:wmt14_bs}
     \end{subfigure}
     \begin{subfigure}[c]{0.2375\textwidth}
         \centering
         \includegraphics[width=.88\textwidth]{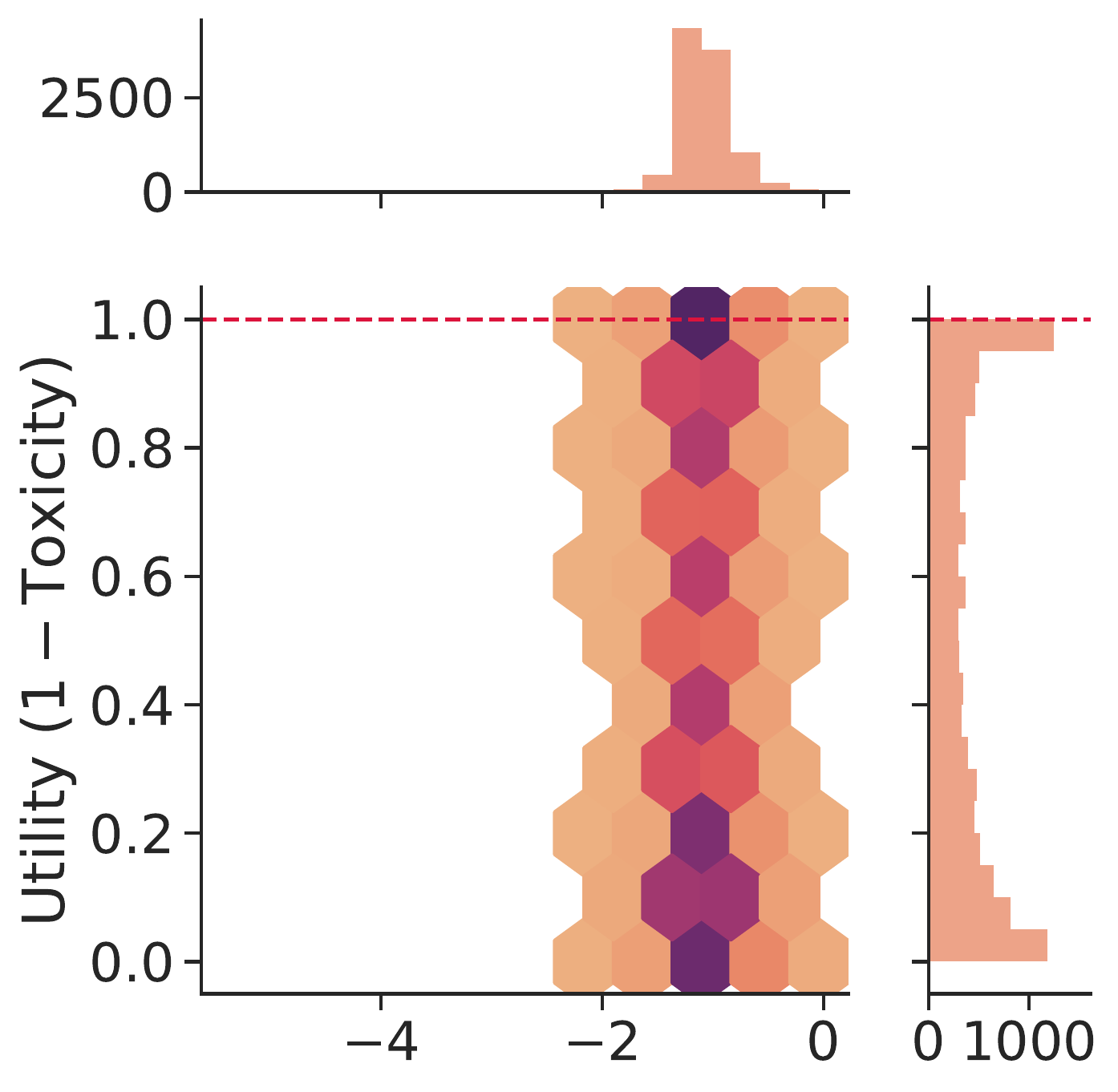}
         \label{fig:rtp_bs}
     \end{subfigure}
     \begin{subfigure}[c]{0.2375\textwidth}
         \centering
         \includegraphics[width=.88\textwidth]{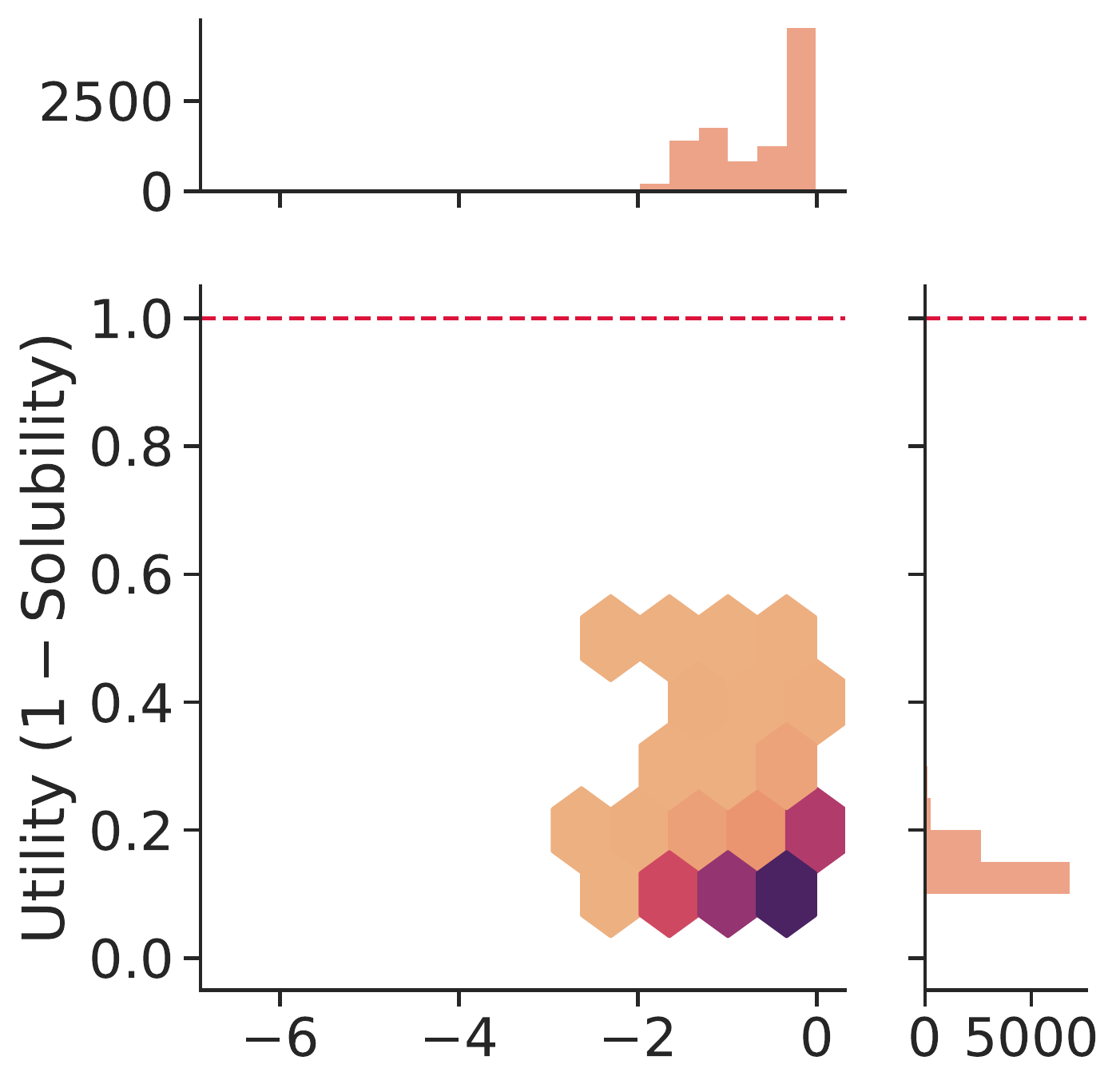}
         \label{fig:prot_bs}
     \end{subfigure}
     
    \rotatebox[origin=c]{90}{\footnotesize Stochastic Beams}
     \begin{subfigure}[c]{0.2375\textwidth}
         \centering
         \includegraphics[width=.88\textwidth]{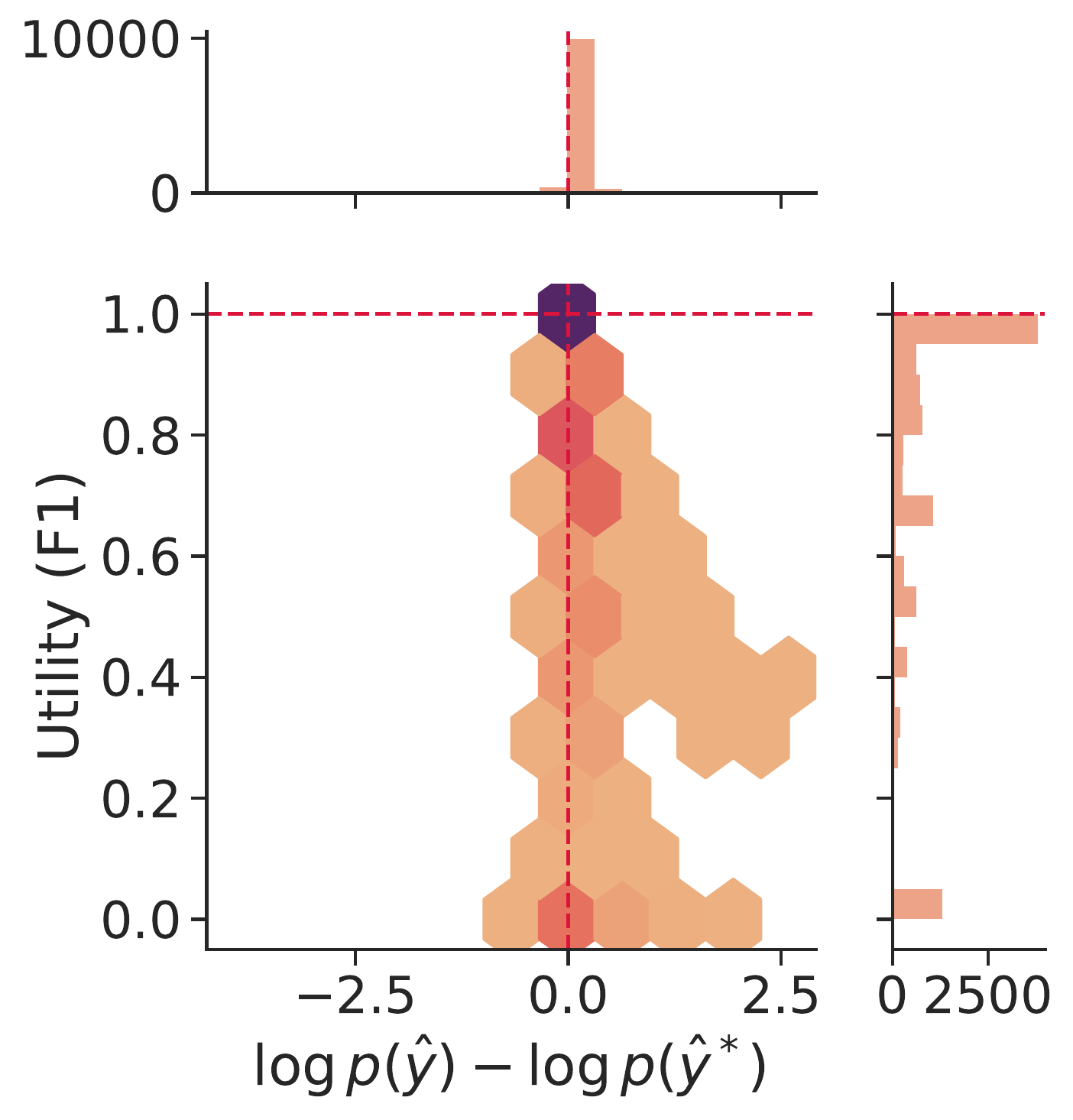}
         \caption{cIE (REBEL)}
         \label{fig:rebel_sbs}
     \end{subfigure}
     \begin{subfigure}[c]{0.2375\textwidth}
         \centering
         \includegraphics[width=.88\textwidth]{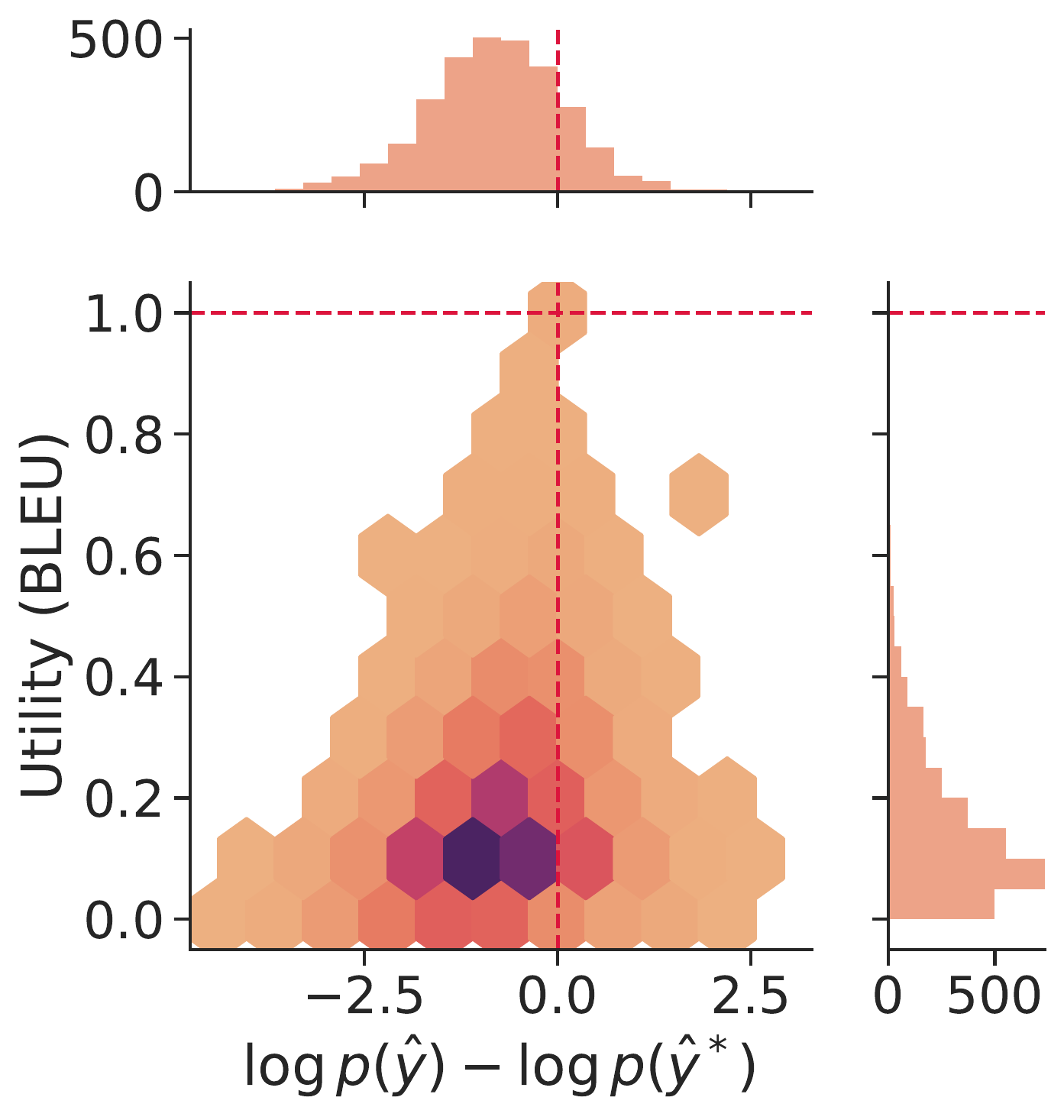}
         \caption{MT (WMT14)}
         \label{fig:wmt14_sbs}
     \end{subfigure}
     \begin{subfigure}[c]{0.2375\textwidth}
         \centering
         \includegraphics[width=.88\textwidth]{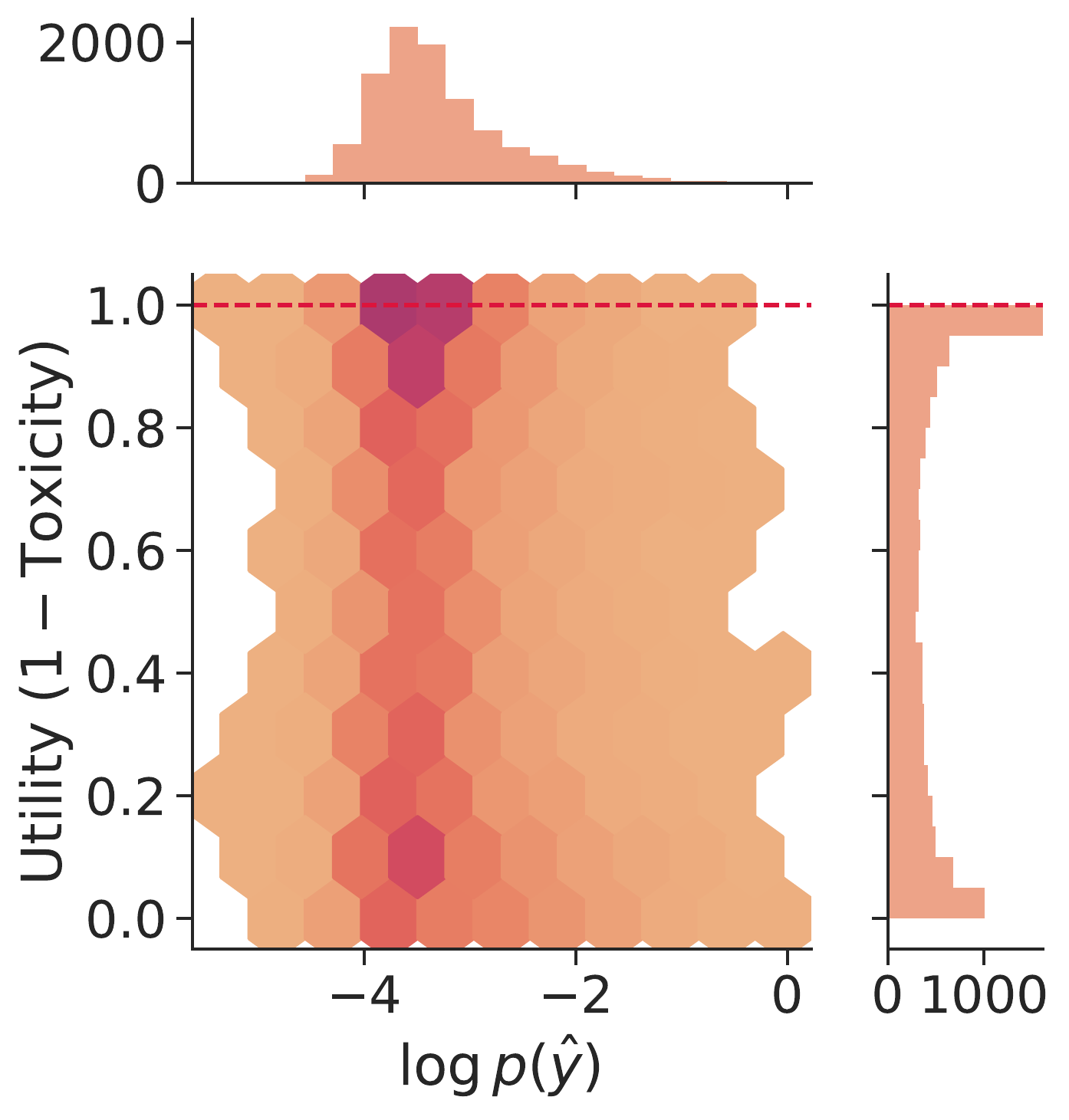}
         \caption{NTTG (RTP)}
         \label{fig:rtp_sbs}
     \end{subfigure}
     \begin{subfigure}[c]{0.2375\textwidth}
         \centering
         \includegraphics[width=.88\textwidth]{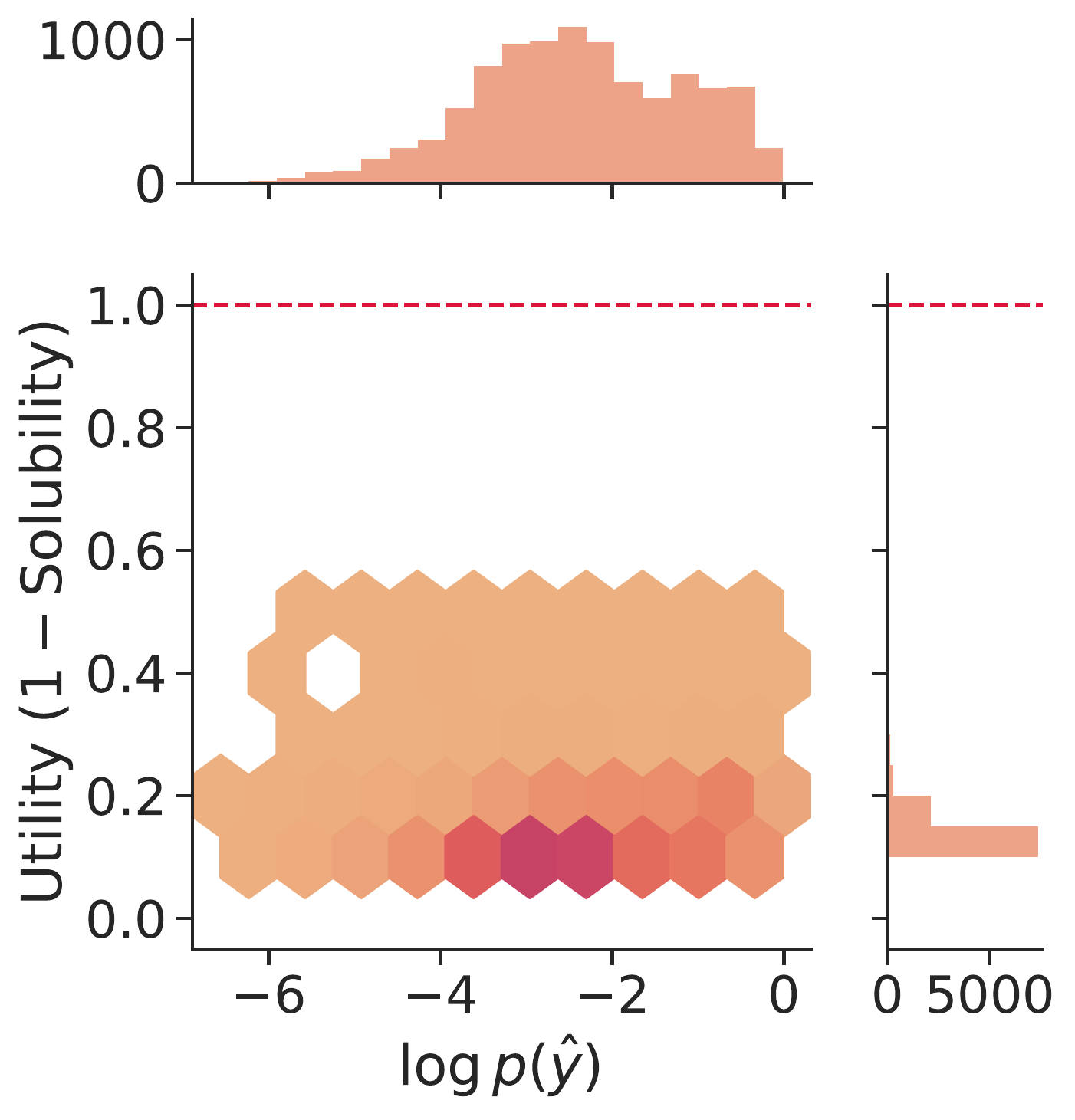}
         \caption{NSPG (SwissProt)}
         \label{fig:prot_sbs}
     \end{subfigure}
     \caption{\textbf{RQ1 (post-decoding alignment)}: Each decoding algorithm is applied to each dataset-model pair. For each subplot, the $x$-axis represents the outputs' log-likelihood under the model, and the $y$-axis the output's task-specific utility score. The plots are frequency heatmaps counting the number of decoded outputs pertaining to a given hexagon. For MT and cIE, the $x$-axis is normalized such that $0$ is the log-likelihood of the target answers{\protect\footnotemark}. These plots show where the outputs are located in the likelihood-utility landscape across tasks and decoding algorithms.}
     \label{fig:fig1}
\end{figure*}

\subsection{Value Models}\label{ssec:value_models}

The quality of a value model 
reflects its ability to approximate
the expected utility. 
To determine the relationship between the value model's quality and the benefit of leveraging it in decoding, we craft models that allow us to instantiate versions with varying levels of quality, ranging from a random predictor to an oracle.

\xhdr{Non-Toxic Text Generation} The state-of-the-art method for detecting toxicity is via classification \cite{Detoxify}. Such a classifier can readily be used as a value model in decoding. We reproduce the training procedure from \citet{Detoxify} and save checkpoints at regular intervals until the training is complete. Due to the gradually decreasing under-fitting, these checkpoints give us a sequence of classifiers that systematically improve in terms of quality.

\xhdr{Machine Translation} To achieve a similar effect for MT, we start by assigning to each data point in the dataset another randomly chosen data point which will serve as a false target. This assignment is fixed across all runs. During inference, the value model calculates the BLEU score for both correct and incorrect targets and returns a linear combination between the points. By gradually increasing the weight assigned to the false target from zero to one, the perfect value model (oracle) slowly degrades to a random predictor.

\section{Experiments and Results}
\label{sec:experiments}

\begin{figure}[t!]
     \centering
     \begin{subfigure}[b]{0.23\textwidth}
         \centering
         \includegraphics[width=0.9\textwidth]{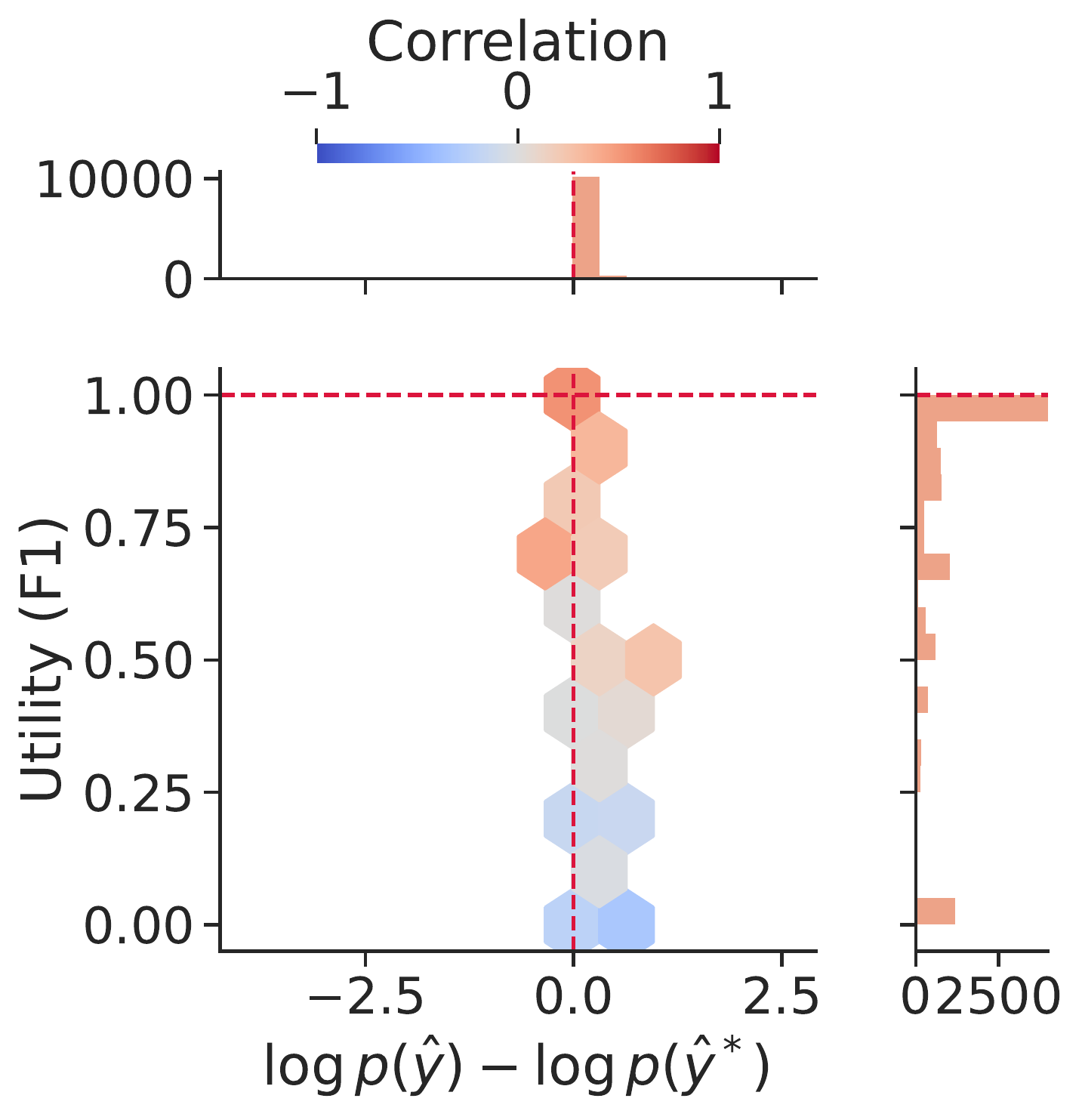}
         \caption{cIE (REBEL)}
         \label{fig:rebel_align_bs}
     \end{subfigure}
     \begin{subfigure}[b]{0.23\textwidth}
         \centering
         \includegraphics[width=0.9\textwidth]{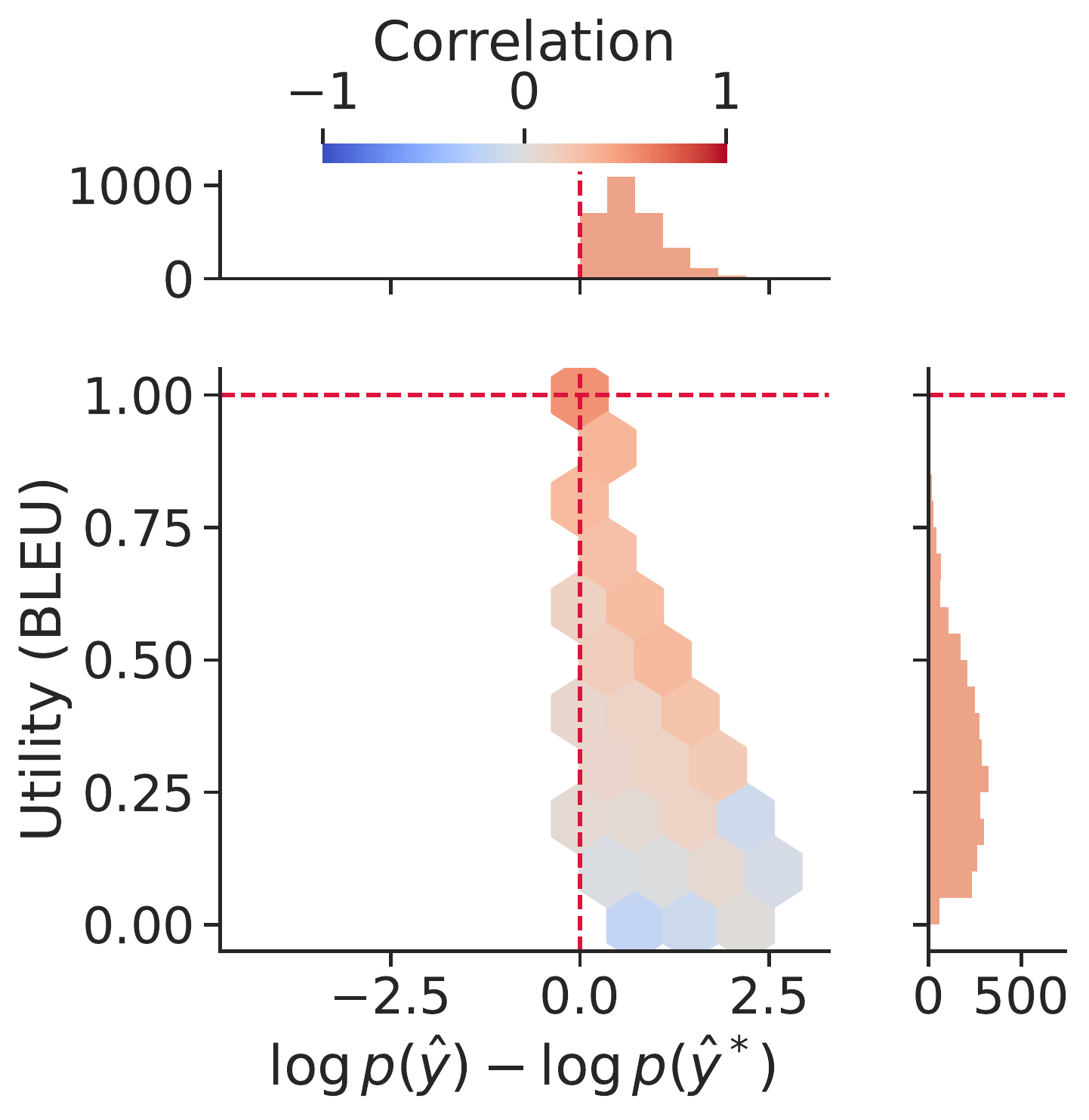}
         \caption{MT (WMT14)}
         \label{fig:wmt14_align_bs}
     \end{subfigure}
     
     \begin{subfigure}[b]{0.23\textwidth}
         \centering
         \includegraphics[width=0.9\textwidth]{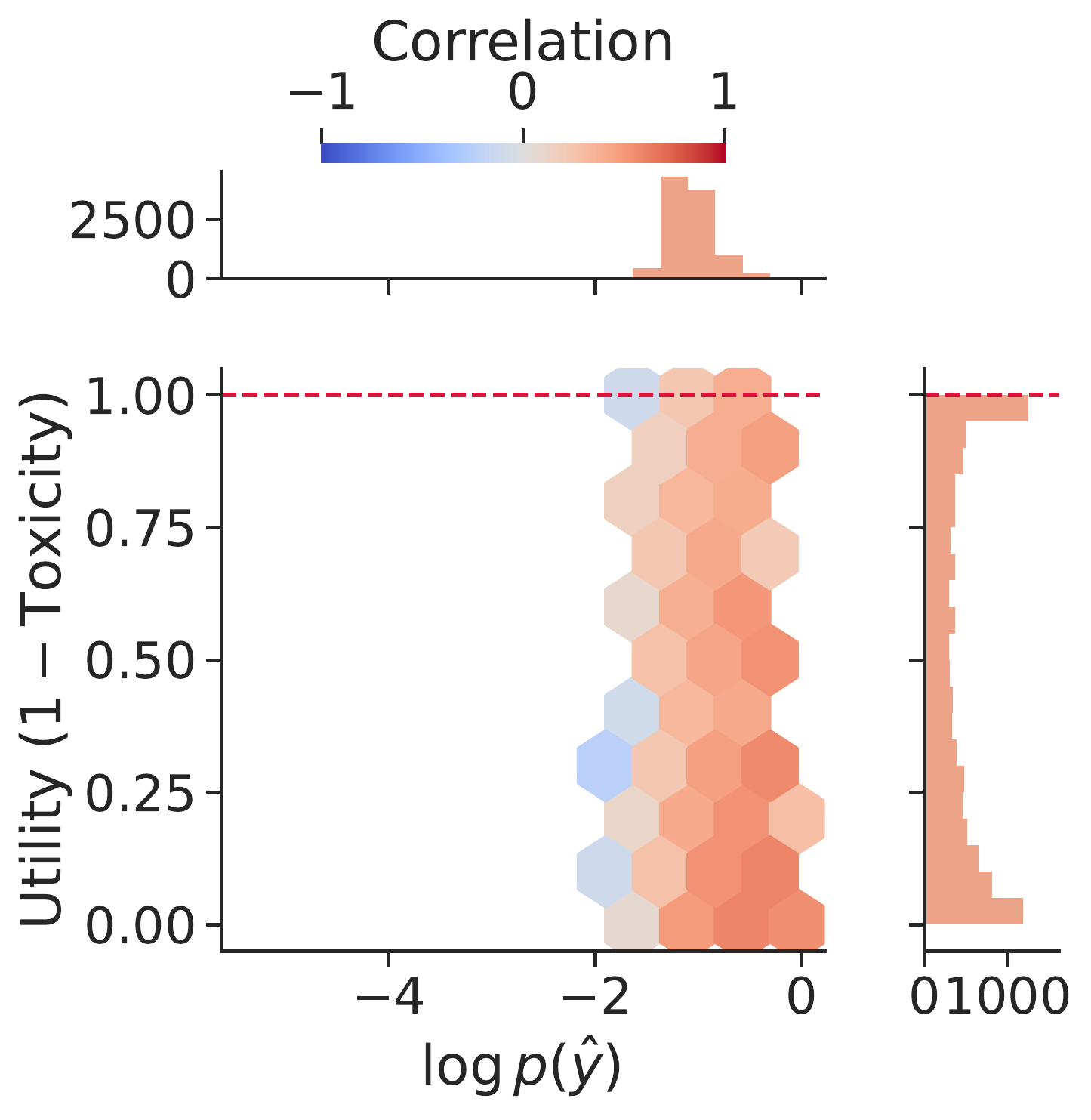}
         \caption{NTTG (RTP)}
         \label{fig:rtp_align_bs}
     \end{subfigure}
     \begin{subfigure}[b]{0.23\textwidth}
         \centering
         \includegraphics[width=0.9\textwidth]{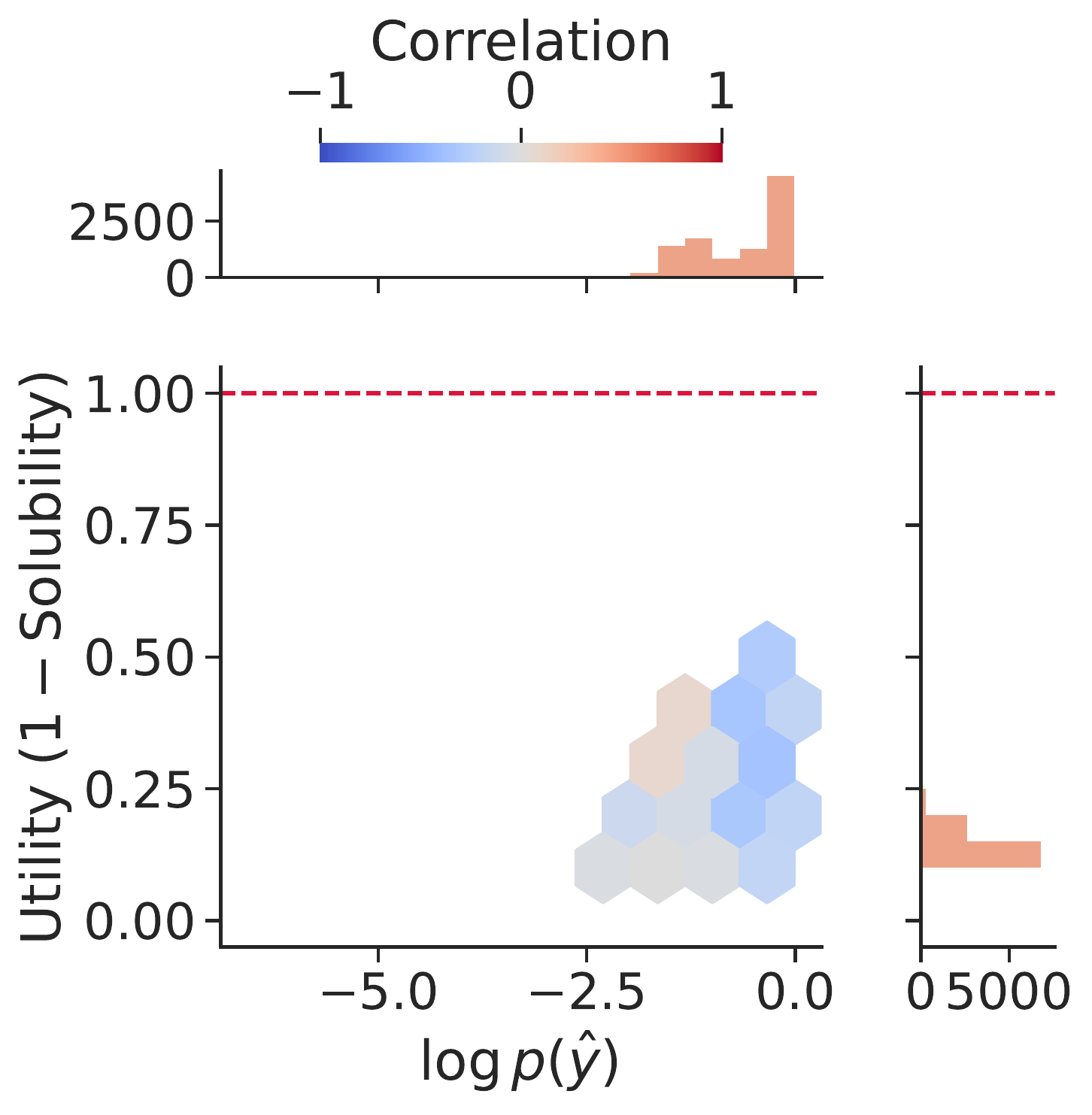}
         \caption{NSPG (SwissProt)}
         \label{fig:prot_align_bs}
     \end{subfigure}

     \caption{\textbf{RQ1 (during-decoding alignment)}: For each dataset-model pair, we run BS and analyze the correlation between likelihood and utility of the top-$5$ candidate hypotheses. 
    The $y$-axis represents the task-specific utility score, and the $x$-axis the log-likelihood under the model. The plots are generated as follows: (i) take the BS outputs from \Figref{fig:fig1} with their log-likelihood and utility scores, which indicate the $x$ and $y$ coordinate of each data point; (ii) for each data point measure the Kendall's $\tau$ correlation between likelihood and utility of the top-$5$ candidate hypotheses; and (iii) average the correlation across the points belonging to the same hexagon.}
     \label{fig:fig2}
     \vspace{-1em}
\end{figure}

\subsection{RQ1: The Likelihood--Utility Relationship}
We present two analyses, one measuring the likelihood--utility misalignment after decoding and one measuring it during decoding.
\footnotetext{See \Figref{fig:fig1_no_target_normalization} for a plot without the target normalization.}

\xhdr{Post-decoding alignment}
In this experiment, we first run each likelihood-based decoding algorithm (GS, BS, and SB) for each dataset--model pair. 
Then, for each output, we compute both the model likelihood and the task-specific utility.\footnote{We also ran experiments with top-$p$ and top-$k$ but did not observe a behavior different from BS.}
We report the results in \Figref{fig:fig1}. 

For cIE (first column), most outputs have
a likelihood close to the targets' likelihood. The majority of the outputs have perfect
utility --- the decoded output is exactly the target, and the model is well-calibrated. This confirms the intuition that when the UD and DS are small, 
greedy likelihood-based MMS are very effective and can cope with the TI.

However, in tasks with larger DS and UD, the story is different. In MT (second column), the combined effect of TI and DS gives rise to a negative global correlation ($-.56$ for BS and $-.52$ for GS in terms of Pearson's correlation; $p < 10^{-3}$)
between the predictions' utility and likelihood after decoding. This is an instance of \emph{Goodhart's law}, where a surrogate metric (likelihood), when being optimized heavily, becomes a poor approximation of the original property it is supposed to track (utility). 

In tasks with large UD, NTTG (third column), and NSPG (fourth column), decoding according to likelihood does not guarantee utility. For example, in NTTG, the likelihood--utility correlation is $-.10$ for GS, .$10$ for SB, and .$03$ for GS in terms of Pearson's correlation; $p < 10^{-3}$.
These scenarios require external information that can guide the decoding towards high-utility outputs.

\xhdr{During-decoding alignment}
Now, we investigate the likelihood--utility alignment where it matters: for outputs close to being extracted by the decoding algorithms. BS maintains $k$ candidate hypotheses, one per beam, before returning the top-scoring one as the final output. In this experiment, we analyze the correlation between the likelihood and utility of the top-$5$ candidates. 
The results are reported in \Figref{fig:fig2}. 
There are three dimensions to this problem: (i) the likelihood (x-axis); (ii) the utility (y-axis); (iii) the correlation (color). Ideally, we would like to see red everywhere, indicating that failure to retrieve a high-utility output is due to the decoding algorithm, but the likelihood of the model is still a good predictor of utility. However, this is not what we observe.

For MT and cIE, we see a clear picture, red color (high likelihood--utility correlation) occurs at the top of the plot (high-utility): 
high likelihood-utility correlation among candidate outputs is enough to yield close to perfect-utility outputs.

For the NTTG (\Figref{fig:rtp_align_bs}), the correlation between utility and likelihood among the beams increases as the likelihood increases. When the model generates high-likelihood outputs, there is a positive correlation between being more likely and being less toxic. However, the likelihood mass is assigned to low-utility regions of the output space, which cannot be resolved with decoding based only on the likelihood. 
For NSPG (\Figref{fig:prot_align_bs}), the correlation across all bins is negative, indicating that high likelihood is a very bad predictor of high utility.

\xhdr{Takeaways}
When TI is the only cause of misalignment, the likelihood is a strong predictor of utility; then, likelihood-based decoding algorithms are expected to retrieve high-utility outputs.
When UD and/or DS are present, the correlation between likelihood and utility post-decoding plummets, indicating that likelihood-based decoding algorithms are ill-suited.
When UD is present (bottom row of \Figref{fig:fig3}), good correlation among the beams does not necessarily mean good utility. However, without UD (top row of \Figref{fig:fig3}), higher correlation among the beams is associated with high utility.

\subsection{RQ2: The Benefits of Value Models}
\label{rq2}
\begin{figure}
     \centering
     \begin{subfigure}[b]{0.23\textwidth}
         \includegraphics[width=\textwidth]{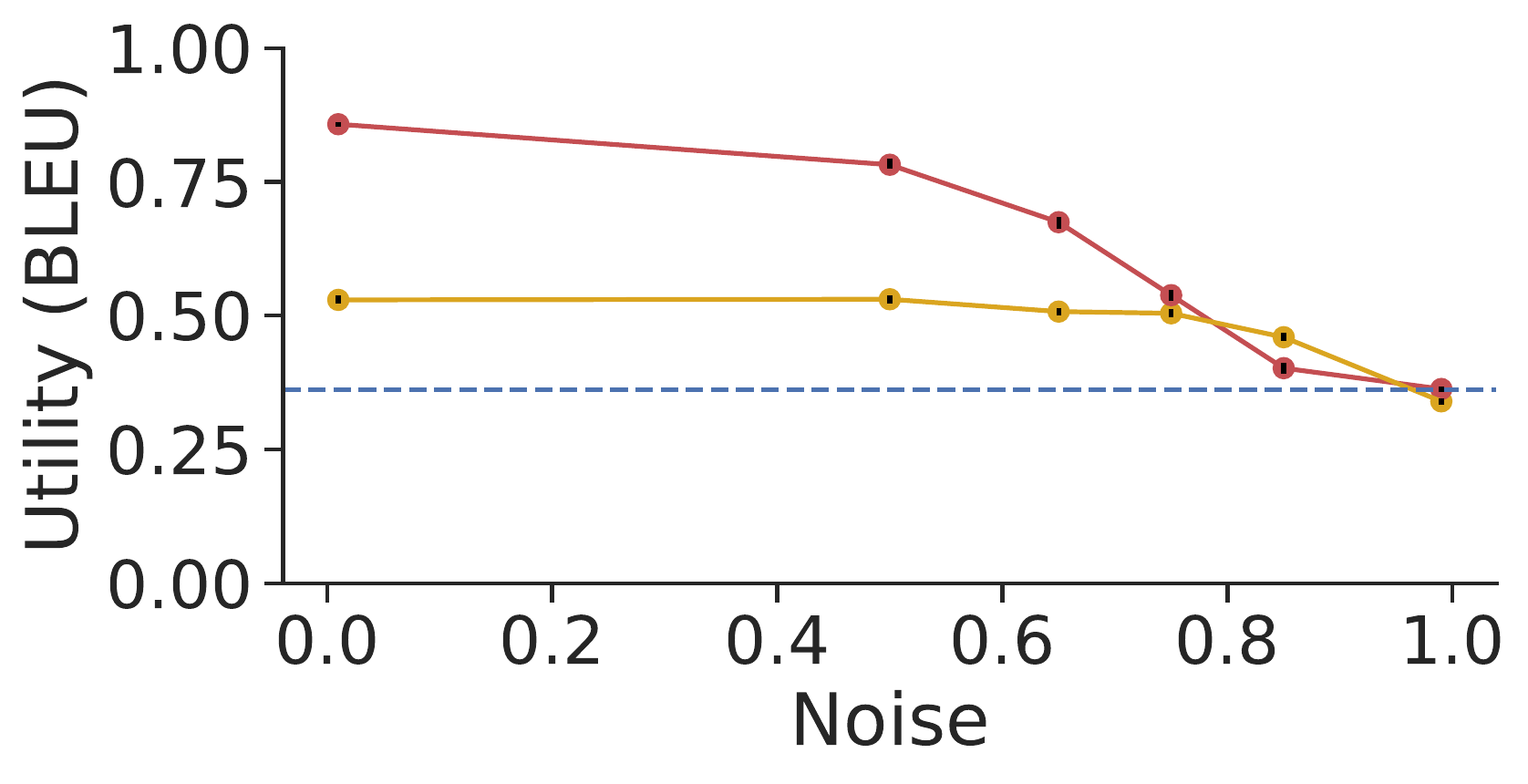}
         \caption{MT (WMT14)}
         \label{fig:wmt14_noise}
     \end{subfigure}
     \begin{subfigure}[b]{0.23\textwidth}
         \includegraphics[width=\textwidth]{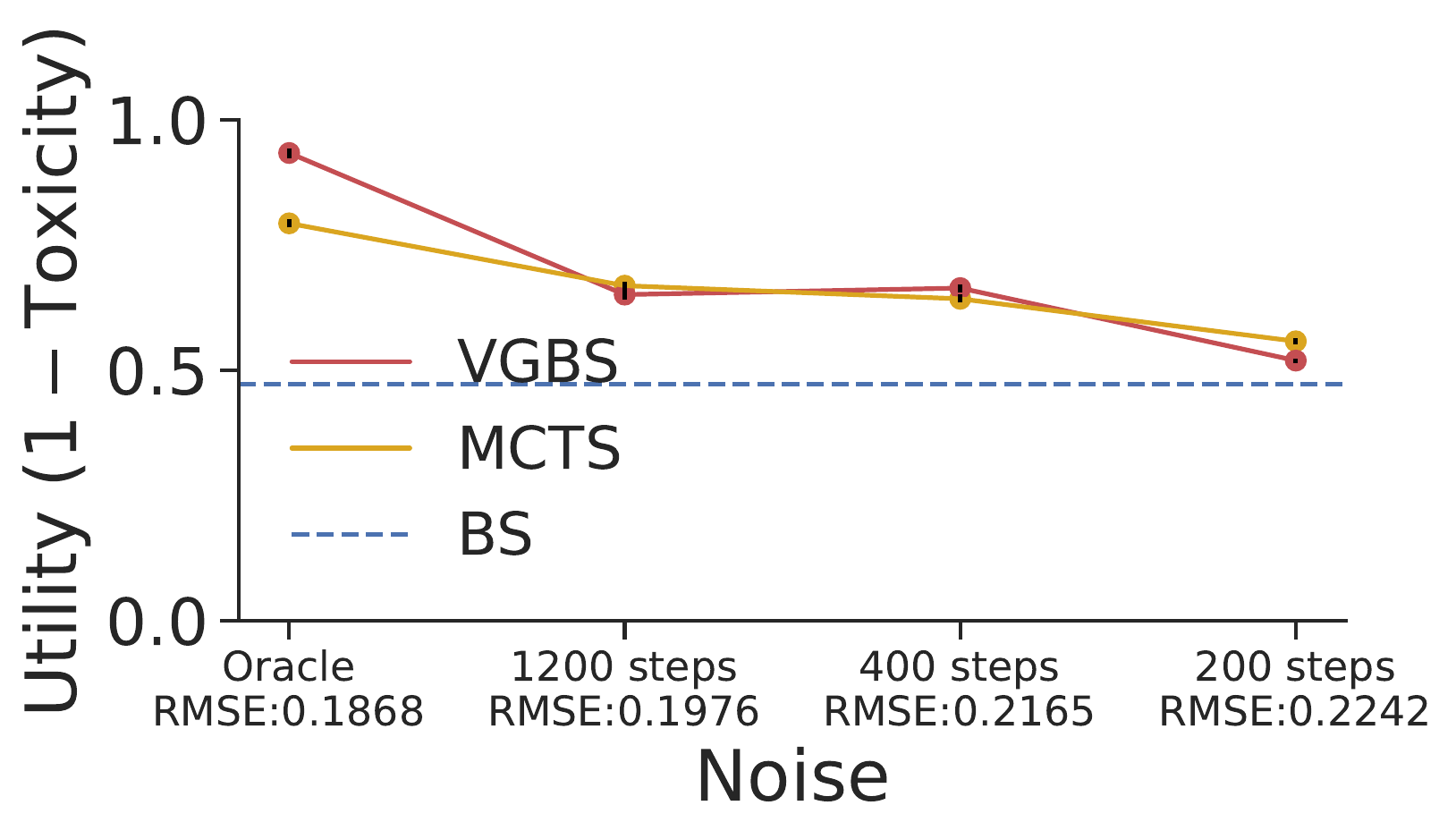}
         \caption{NTTG (RTP)}
         \label{fig:rtp_noise}
     \end{subfigure}
     \caption{\textbf{RQ2}: For MT and NTTG, we ran VGBS and MCTS with value models displaying various levels of noise.
     We report the average utility of outputs on the $y$-axis (with 95\% confidence interval). The noisy value models are described in \Secref{ssec:value_models}.}
     \label{fig:fig3}
\end{figure}
We now analyze the benefits of value-guided decoding algorithms (VGBS and MCTS) as a function of the value model's quality (see \Secref{ssec:value_models}).  
Due to the high computational cost of running the experiments with both VGBS and MCTS, we focus on two tasks: MT and NTTG. For each version of the value model, we first perform a hyperparameter search on a small subset of the data and use the best hyperparameters on the test set. The results are reported in \Figref{fig:fig3}.
VGBS and MCTS are always at least as good as BS, even with random value models, as the small-scale hyperparameter search selects parameters that ignore the values when they are not useful. 
However, when there is some signal in the value model, both VGBS and MCTS effectively leverage it and quickly start outperforming BS. When the value model is accurate, very high-utility outputs are discovered. Interestingly, VGBS mostly outperforms MCTS, and  
can extract almost perfect outputs, whereas MCTS plateaus. 
This is significant because VGBS has a substantially lower complexity than MCTS (see \Appref{app:complexity}). 

\xhdr{Takeaways}
Value-guided decoding algorithms can overcome the likelihood--utility misalignment and significantly outperform likelihood-based decoding algorithms even with noisy value models, as long as a small-scale hyperparameter search is done. 
VGBS offers a better trade-off between performance and computation cost than MCTS.

\subsection{RQ3: Prompting as an MMS}
\begin{figure}[t!]
     \centering
     \begin{subfigure}[b]{0.35\textwidth}
        \includegraphics[width=0.9\textwidth]{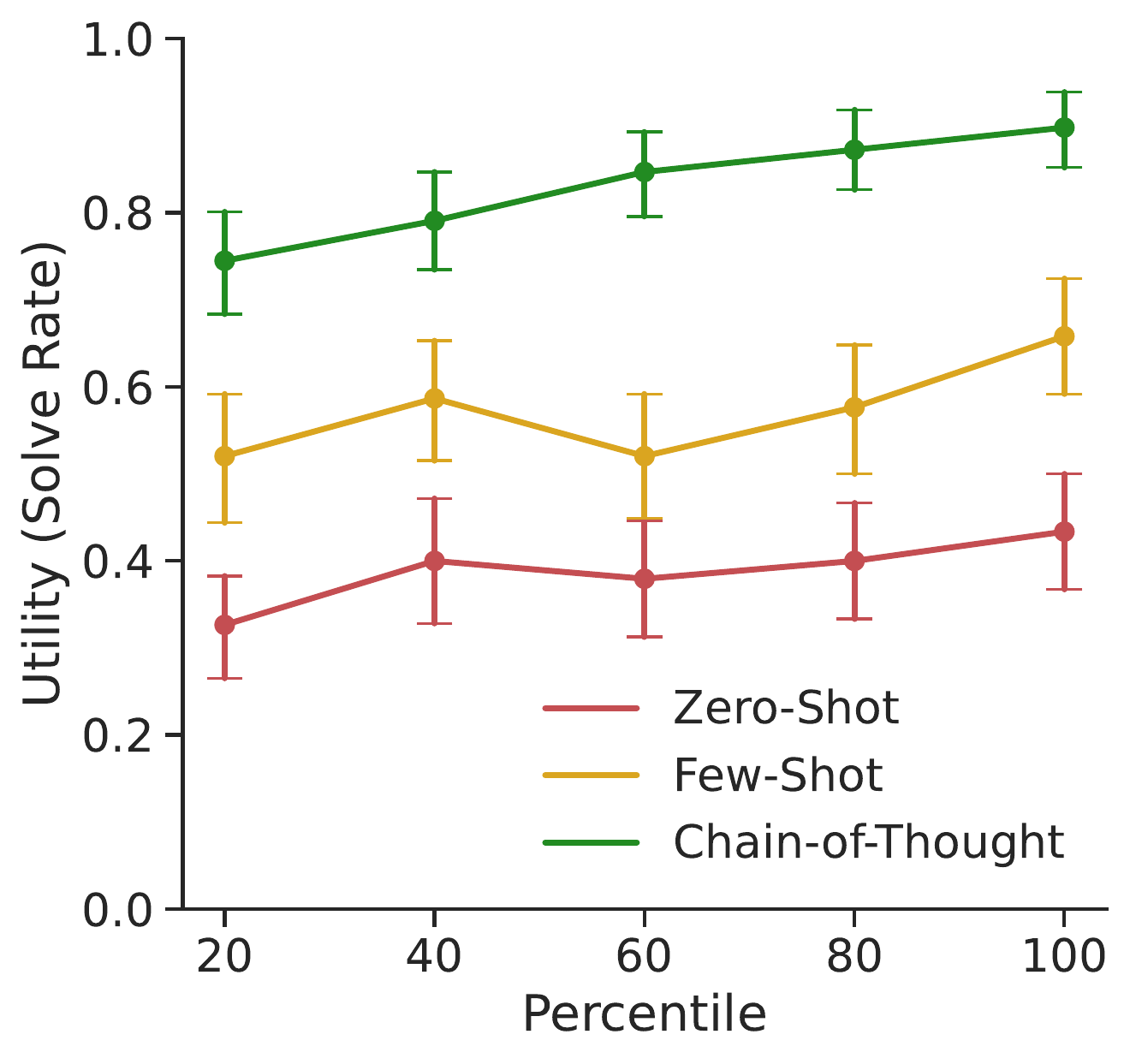}
     \end{subfigure}
     \caption{\textbf{RQ3}: ZS, FS, and CoT prompting, on the Sports understanding dataset. We report the utility ($y$-axis) of outputs binned according to the empirical percentiles of their likelihood ($x$-axis). The lineplot is the average utility per bin with 95\% confidence intervals.
     }
     \label{fig:fig5}
\end{figure}
Recently, \citet{DBLP:journals/corr/abs-2201-11903} showed how the simple and broadly applicable idea of including in the prompt a few examples where the targets contain a sequence of steps that lead to the answer can greatly enhance the reasoning capabilities of LMs. We mimic the Sports Understanding task in their work by taking the same in-context examples and evaluating the same two prompting methods: CoT and standard FS prompting. Additionally, we evaluate the model in the standard zero-shot (ZS) setting, without any examples in the context, as a baseline. This results in unstructured answers that need to be labeled manually, see \Appref{app:exp_and_results_rq3}.

To test our hypothesis that prompting is a means of addressing misalignment, we measure the utility and the likelihood under the model for all the testing data points. 
The results, summarized in Fig.~\ref{fig:fig5}, provide three insights. First, similarly to \citet{DBLP:journals/corr/abs-2201-11903}, 
CoT outperforms FS and ZS with an accuracy of 83\%, versus 57.3\% and 38.9\% for FS and ZS, respectively.
Second, (and complementary to the information visualized on the plot) the average log-likelihood of the outputs generated by CoT is significantly higher than the FS and ZS generated outputs, $-0.067$ as opposed to $-0.17$ and $-2.472$. Third, the correlation between the likelihood and the utility when decoding with CoT is higher: 0.11 Pearson's correlation compared to 0.07 and 0.09 for FS and ZS, respectively.%
\footnote{The differences are statistically significant ($p < 10^{-3}$).}
Referring back to the observation made in RQ1b on \Figref{fig:fig2}, the value-guided MMSs studied in \Secref{rq2} address the misalignment post-hoc. However, the hidden representations building up to that misalignment are not modified, and the undesired information will still be attended to in predicting the next token probability distribution. In contrast, an effective prompting strategy addresses the misalignment before it affects the hidden representations, thereby (i) forcing the model to assign high likelihood to high-utility regions of the output space and (ii) improving the likelihood--utility alignment, making it easier to find high-utility outputs with greedy likelihood-based decoding algorithms.

\xhdr{Takeaways}
Effective prompting methods put the model in a state where the generated outputs' likelihood is well-aligned with the desired utility.

\section{Discussion}
\label{sec:discussion}

RQ1 reveals that decoding based solely on the likelihood gives poor expected utility whenever DS or UD occurs. DS and UD make the likelihood a poor predictor of utility.
When only TI is present, these decoding algorithms perform well because the likelihood is a strong predictor of utility.

Then, in RQ2 and RQ3, we saw that methods bringing in external information at decoding time manage to effectively solve the likelihood--utility misalignment problem. 
While finetuning (or retraining) would be an obvious and apparently ideal MMS, 
this is often neither possible nor necessary.
Indeed, our experiments show that if a value model can be crafted and we can afford the extra compute for the value model calls, then VGBS becomes a strong decoding algorithm capable of fixing misalignment problems at inference time. It is more efficient than MCTS and performs better than BS, even if the value model is only a poor approximator of the utility. When crafting a useful value model is difficult (e.g., protein function depends on the 3D structure, which cannot be easily approximated from partial amino-acid sequences), MCTS with a large number of simulations with roll-outs can be used to ``estimate'' one. However, the price to pay is a higher computational cost at inference time. Finally, for large, generalist LMs, decoding algorithms such as MCTS or VGBS are prohibitively expensive due to the high computational cost of each call to the LM. Prompting methods combined with greedy or top-$p$ decoding can be considered as a way to leverage external information in the form of few-shots prompts to set the model in a state where the likelihood is better aligned with the utility. Our experiments support this explanation of the success of prompting large LMs. 
For a similar perspective, comparing prompting methods with training-based MMSs 
see \newcite{DBLP:journals/corr/abs-2110-04366}.

\section*{Limitations}
\label{sec:limitations}

\xhdr{Non-exhaustive empirical analysis} This work studies a fundamental problem that involves a complex interaction between tasks, models, and data; and sequence-to-sequence models have been applied to a very broad set of tasks. Covering all possible combinations is impossible, and for our empirical analysis, we chose a subset to evaluate. Our choice is guided by the classification of misalignment sources proposed in \Secref{sec:4.2_task_datasets} and aims to cover different areas of the misalignment space. 
A seemingly small difference between two choices (e.g., a difference in the loss function used in training the model) can give rise to a considerably different misalignment and, consequently, performance.
This is why the goal of the proposed conceptual framework is to make a step toward enabling a more systematic study of decoding. To further help the community investigate the broader space of tasks, models, and datasets through this lens, we open-source the implementation of our analysis.

\xhdr{Alternative ways of fixing misalignment} Apart from value-based decoding, other techniques could be considered to fix the misalignment problem:
(a) Retrain or finetune the model with data that better reflects the task's utility. For instance, to generate non-toxic text, one could retrain or finetune GPT2 on curated datasets that contain toxic prompts and non-toxic sentence continuations. 
(b) Optimize more directly the utility function instead of surrogate differentiable objectives. This could be done via reinforcement learning (see \citet{wang-etal-2018-deep, wu-etal-2018-study} for BLEU).

In this work, we focused on decoding algorithms and ways of fixing the likelihood--utility misalignment problem at inference time. Future research could further investigate the trade-offs involved in finetuning and retraining. Is it better to invest resources in acquiring new data that fits the task for finetuning? Or is it better to fix DS and UD at inference time with VGBS, MCTS or prompt engineering? Where do the inflection points lie?

\section*{Acknowledgments}
This work was conducted in the context of the Microsoft Turing Academic Program (MS-TAP).
West's lab is partly supported by grants from
Swiss National Science Foundation (200021\_185043),
Swiss Data Science Center (P22\_08),
H2020 (952215),
Microsoft Swiss Joint Research Center,
and Google,
and by generous gifts from Facebook, Google, and Microsoft.

\bibliography{anthology,custom}
\bibliographystyle{acl_natbib}

\appendix
\clearpage

\section{Proposed MMS Taxonomy}
This section describes some of the most prominent members in each class of the proposed taxonomy.

\subsection{Greedy Likelihood-Based Strategy}
\label{appendix:taxonomy_greedy}

\subsubsection{Deterministic}

\xhdr{Greedy Search (GS)} The simplest among the decoding algorithms, at each step \(t\), GS selects the token with the highest likelihood under the model.

\xhdr{Beam Search (BS)} An extension of GS, BS, maintains not one, but \(k \in \mathbb{N}^{+}\) partially-decoded sequences, called beams, in parallel. At each step \(t\), BS: (i) pre-selects the most likely $k$ tokens for each beam; (ii) from the resulting $k \times k$ nodes, the algorithm selects the \(k\) with the highest likelihood and drops the rest.

\subsubsection{Stochastic}

An alternative that increases the diversity of output sequences is to sample the tokens at each step from the likelihood distribution $\hat{y}_t \sim p(y_t|\hat{y}_{<t}, x)$. Instead of sampling from the full distribution, these decoding algorithms typically focus greedily on the tokens corresponding to the high-probability regions.

\xhdr{Top-$k$ Sampling} Top-$k$ samples the next tokens from a truncated distribution where only the $k$ most probable tokens are considered \citep{fan-etal-2018-hierarchical}.

\xhdr{Top-$p$ Sampling} Top-$p$ samples the next tokens from a truncated distribution where only the smallest set of tokens with a probability mass bigger than (or equal to) $p$ is considered \citep{DBLP:conf/iclr/HoltzmanBDFC20}.

\xhdr{Stochastic Beams (SB)} SB samples completed outputs without replacements according to the LM's likelihood. The implementation relies on applying BS on likelihood scores perturbated with Gumbel noise \cite{pmlr-v97-kool19a}.

\subsection{Greedy Likelihood-Based Strategy with Pruning}
\label{appendix:taxonomy_greedy_with_pruning}

\xhdr{Ad-Hoc Heuristics} Currently, most tasks utilize some ad-hoc heuristics. 
For instance, in MT, it is often necessary to discourage empty (or short) sequences by enforcing a minimal sequence length \cite{stahlberg-byrne-2019-nmt}.
Similarly,  state-of-the-art language generation models often get stuck in repetitive loops. Therefore, an $n$-gram repetition penalty is now part of the standard toolkit \citep{klein-etal-2017-opennmt}.

\xhdr{Constrained Beam Search (CBS)} 
The idea of constraining the likelihood during decoding can be extended to include task-specific knowledge. For example, in information extraction tasks, the BS decoding strategy has been constrained to only extract outputs satisfying the predefined schema \cite{scholak-etal-2021-picard, de-cao-etal-2022-multilingual, josifoski-etal-2022-genie}. Then, BS only searches high-scoring outputs among the smaller subset of valid ones.

\xhdr{NeuroLogic} The NeuroLogic strategy enforces the satisfaction of given lexical constraints by controlling the decoding stage of sequence generation \citep{lu-etal-2022-neurologic}. While BS aims to maximize the likelihood of the generated sequence, NeuroLogic searches for optimal output sequences among the strings that also satisfy the given constraints. Hard logic constraints are converted into a soft penalty term in the decoding objective, and beam-based search is used to find approximately optimal solutions.

\subsection{Greedy Likelihood- and Value–Based Strategy}
\label{appendix:taxonomy_value_based}

\xhdr{Value-Guided Beam Search (VGBS)} It is the most intuitive example of a greedy decoding algorithm that leverages a value model \cite{NIPS2017_2b24d495, DBLP:conf/cvpr/RenWZLL17}. It uses a greedy strategy similar to BS but selects the next token using a linear combination of the LM's likelihood and the scores from the value model.

More specifically, instead of expanding each beam by the \(m\) highest-scored tokens according to the likelihood, the algorithm chooses the top \(m\) tokens according to the following scoring function:
\begin{equation*}
    s_{y_{<i}, x}(y_i) = \frac{\alpha}{i}\log(p(y_{<i}y_i|x)) + (1-\alpha)v(y_{<i}y_i, x),
\end{equation*}
where the factor \(\alpha\) weights the contribution of the the value model, $y_{<i}$ denotes the partially decoded sequence, and $y_i$ corresponds to the next token under consideration.

\subsection{Simulation-Based Strategy}
\label{appendix:taxonomy_simulation_based}
\xhdr{Monte\hyp Carlo Tree Search (MCTS)}
MCTS is the canonical example of simulation-based tree exploration informed by value. In our setup, it differs from all other decoding algorithms because, at step $i$, it may explore sequences of length greater than $i$. It is not tied to committing to local decisions without exploring the tree. In each step, MCTS has a fixed computational budget that it uses to explore multiple paths before choosing the next token. 
For more details, we refer to \newcite{chaffin-etal-2022-ppl}, whose implementation we adapt for this work.

\subsection{Prompting-Based Strategy}
\label{appendix:taxonomy_prompting}

\xhdr{Few-Shot (FS)} At inference time, instead of only passing the input $x$, a context comprised of $k$ examples $(x_i, y_i)_{i=1}^k$ is added as a prefix. The main idea is that the model will build on its semantic understanding of the relation between $x_i$ and $y_i$ and make the ``guided'' likelihood better aligned with the utility \cite{NEURIPS2020_1457c0d6}.

\xhdr{Chain-of-Thought (CoT)} The CoT decoding method \cite{DBLP:journals/corr/abs-2201-11903} is a conceptual extension of FS which presents the examples' targets as a sequence of steps that lead to the solution. This format is particularly helpful for tasks that require multi-step reasoning, with which transformers generally struggle.

\section{Experimental Setup}
This section provides additional details about the experimental setup.

\subsection{Details about Data, Models, and Utility Functions}
\label{app:data}

In \Tabref{tab:data_description}, we present a summary of the tasks, utility functions, misalignment types, model, and dataset. We now give a brief description of each task:

\textit{Closed Information Extraction} (cIE) with the REBEL dataset \cite{huguet-cabot-navigli-2021-rebel-relation}
and GenIE model \cite{josifoski-etal-2022-genie} (an instance of BART finetuned to extract the exhaustive set of triples in a sentence following the Wikidata schema). The utility is the F1 score between the generated and the target set of triples. 

\textit{Machine Translation} (MT) with the WMT14 dataset \cite{bojar-EtAl:2014:W14-33} and a pretrained mBART50 model \cite{tang-etal-2021-multilingual} to translate \textit{English to French}. The notion of utility is the match between the generated and the target translation, as measured by BLEU-4.

\textit{Non-Toxic Text Generation} based on the Real Toxicity Prompt (RTP) dataset \citep{gehman-etal-2020-realtoxicityprompts} for prompting a GPT2 model. The notion of utility is whether the generated output contains toxic language or not. The utility function is an ALBERT model \citep{Detoxify} trained on the Jigsaw dataset with an unintended bias to measure the toxicity of a text. 

\textit{Non-Soluble Protein Generation}: We use the SwissProt-EF dataset \cite{swissprot} for prompting a ProtoGPT2 model \citep{Ferruz2022ProtGPT2IA}, which is pretrained on sequences of amino acids from protein prompts. The notion of utility is whether the generated protein is \textit{soluble} or not. To measure non-solubility, we use ProtBERT \cite{protbert}, which is a BERT-based model trained on a large corpus of protein sequences in a self-supervised fashion.  Finally, 

\textit{Sports Understanding} with the Sports Understanding (SU) task, part of the BIG-bench effort \cite{bigbench}, with a 530B parameter pre-trained language model: MT-NLG \citep{tnlgv2}. The primary purpose of this task is to test the general understanding of sports by asking the model to discriminate between plausible and implausible statements relating to sports. 

\begin{table}
\centering
\resizebox{0.9\columnwidth}{!}{%
\begin{tabular}{@{}r|c|c@{}}
\toprule
        & \textbf{LM calls}          & \textbf{Value calls}        \\ \midrule
Greedy Search    & N                 & --                        \\
Beam Search      & N $\times$ B      & --                      \\
Stochastic Beams & N $\times$ B      & --                       \\
VGBS             & N $\times$ B      & N $\times$ B $\times$ K  \\
MCTS             & N $\times$ S      & N $\times$ S             \\ \bottomrule
\end{tabular}%
}
\caption{Coarse complexity analysis of the decoding algorithms used, in terms of LM calls and Value calls. N is the number of tokens to be generated, B the number of beams, K the number of next tokens considered by the value model per beam in VGBS, S the number of simulations per generated token in MCTS. In all our experiments, B=$5$, K=$20$, S=$50$.}
\label{tab:complexities}
\end{table}

\subsection{Hyperparameters of Decoding Algorithms}
\label{app:hyperparam}
The number of beams for BS, SB, VGBS is fixed to 5 for all tasks, except for cIE where it is 10 --- the model's default; and the number of simulations in MCTS is fixed to 50. Due to the high computational cost, to decide the optimal value for MCTS's $c_{puct}$ and VGBS's $\alpha$ in RQ3, we run a hyperparameter search for each level of noise over a small sample of 80 data points (see \Appref{app:compute_and_runtime} for the ranges of the search). For both of the prompting-based strategies, we use greedy decoding during inference.

\begin{figure*}[ht!]
     \centering
     \rotatebox[origin=c]{90}{\footnotesize Greedy Search}
     \begin{subfigure}[c]{0.2375\textwidth}
         \centering
         \includegraphics[width=.88\textwidth]{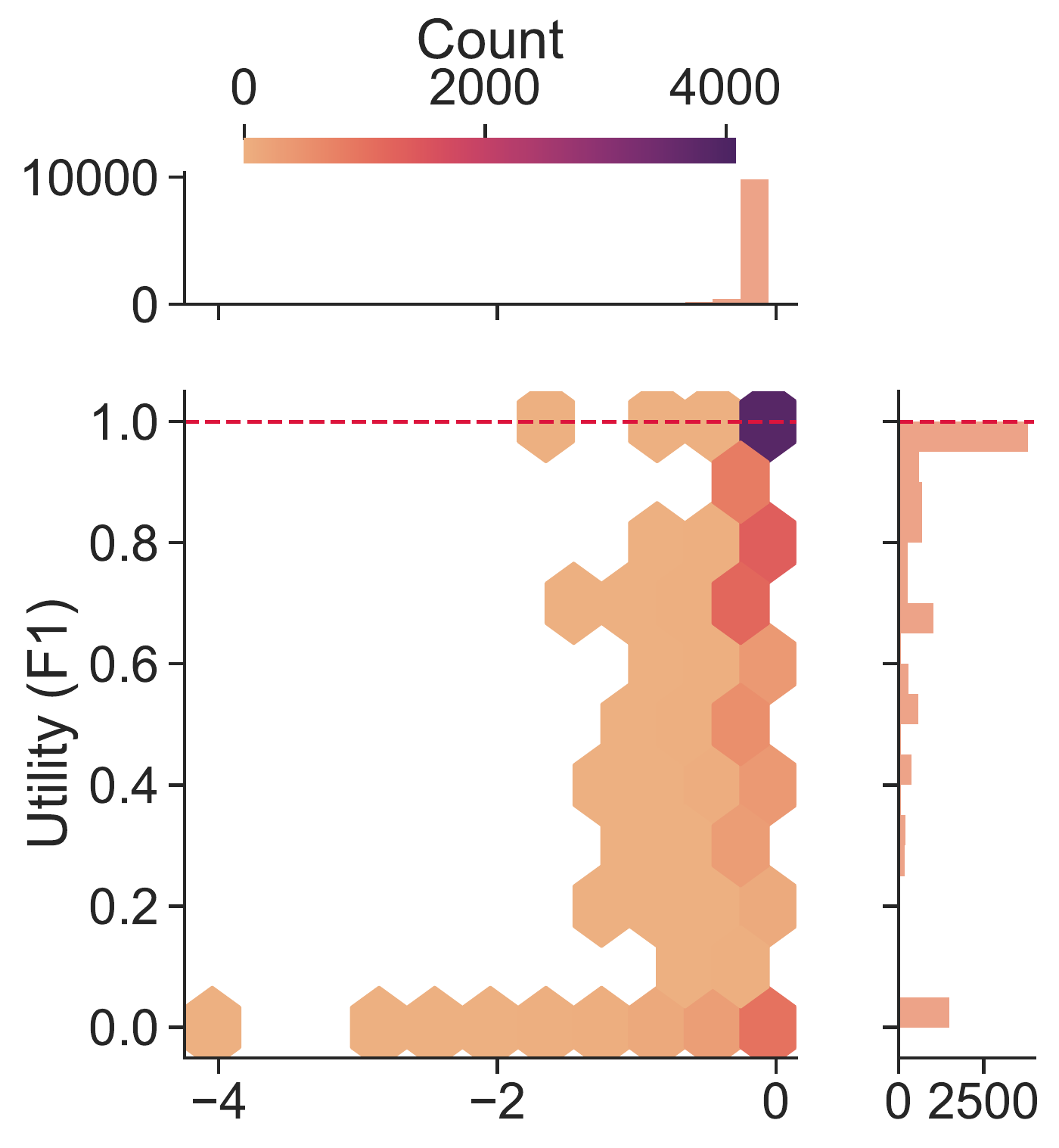}
         \label{fig:rebel_greedy_no_target_normalization}
     \end{subfigure}
     \begin{subfigure}[c]{0.2375\textwidth}
         \centering
         \includegraphics[width=.88\textwidth]{images/fig1_v3/Translation_greedy_search.pdf}
         \label{fig:wmt14_greedy_no_target_normalization}
     \end{subfigure}
     \begin{subfigure}[c]{0.2375\textwidth}
         \centering
         \includegraphics[width=.88\textwidth]{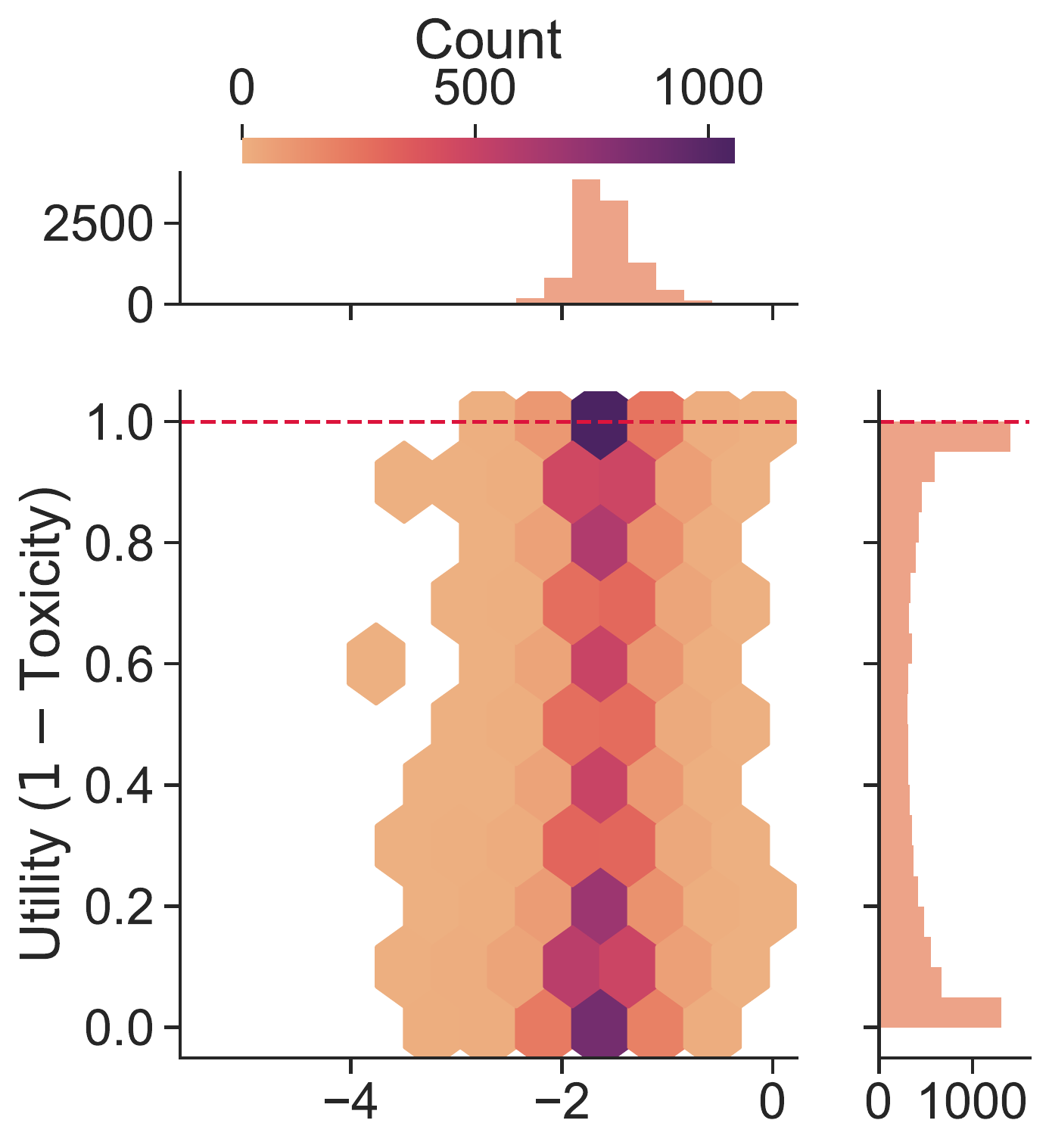}
         \label{fig:rtp_greedy_no_target_normalization}
     \end{subfigure}
     \begin{subfigure}[c]{0.2375\textwidth}
         \centering
         \includegraphics[width=.88\textwidth]{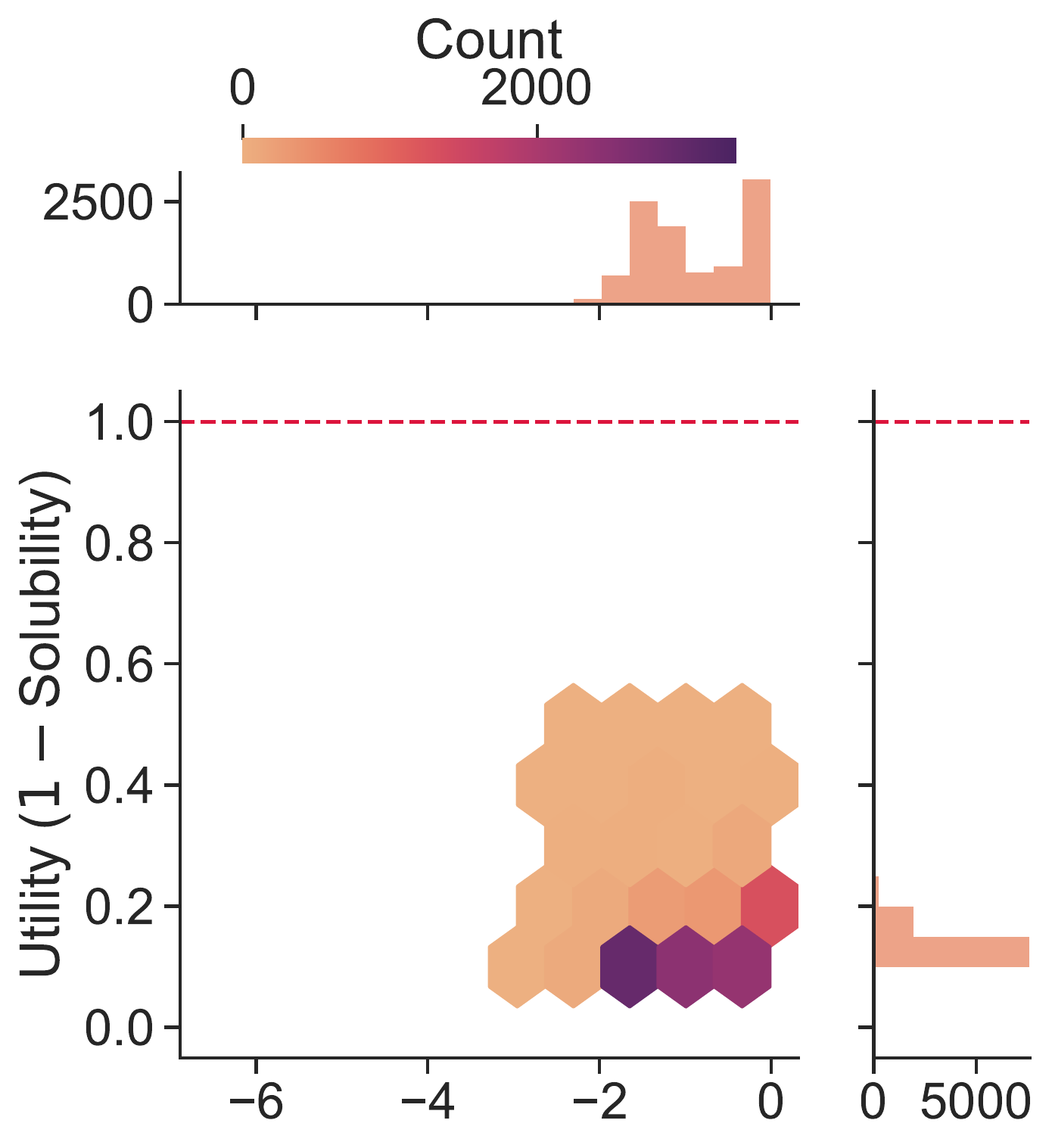}
         \label{fig:prot_greedy_no_target_normalization}
     \end{subfigure}
    \rotatebox[origin=c]{90}{\footnotesize Beam Search}
     \begin{subfigure}[c]{0.2375\textwidth}
         \centering
         \includegraphics[width=.88\textwidth]{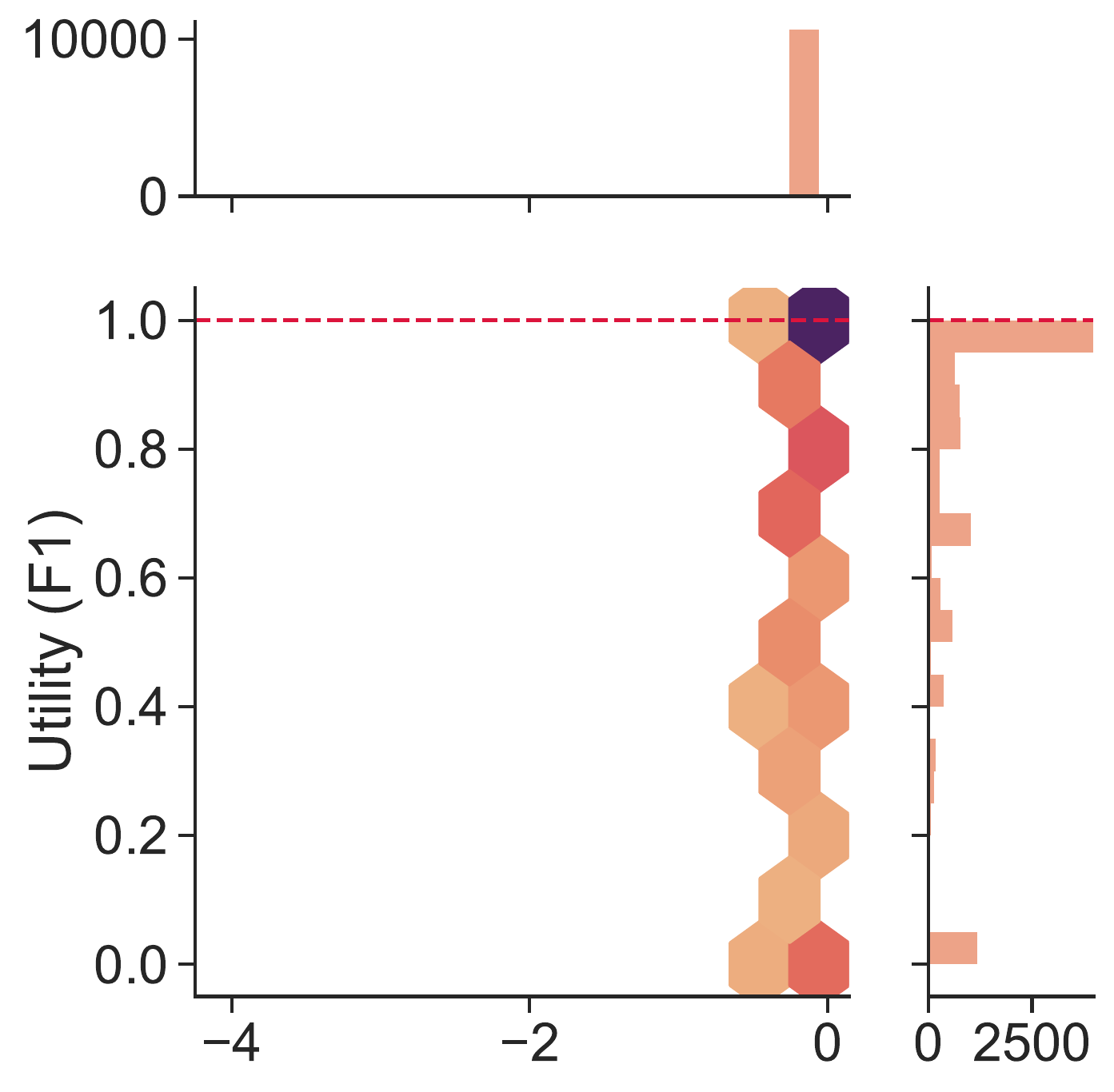}
         \label{fig:rebel_bs_no_target_normalization}
     \end{subfigure}
     \begin{subfigure}[c]{0.2375\textwidth}
         \centering
         \includegraphics[width=.88\textwidth]{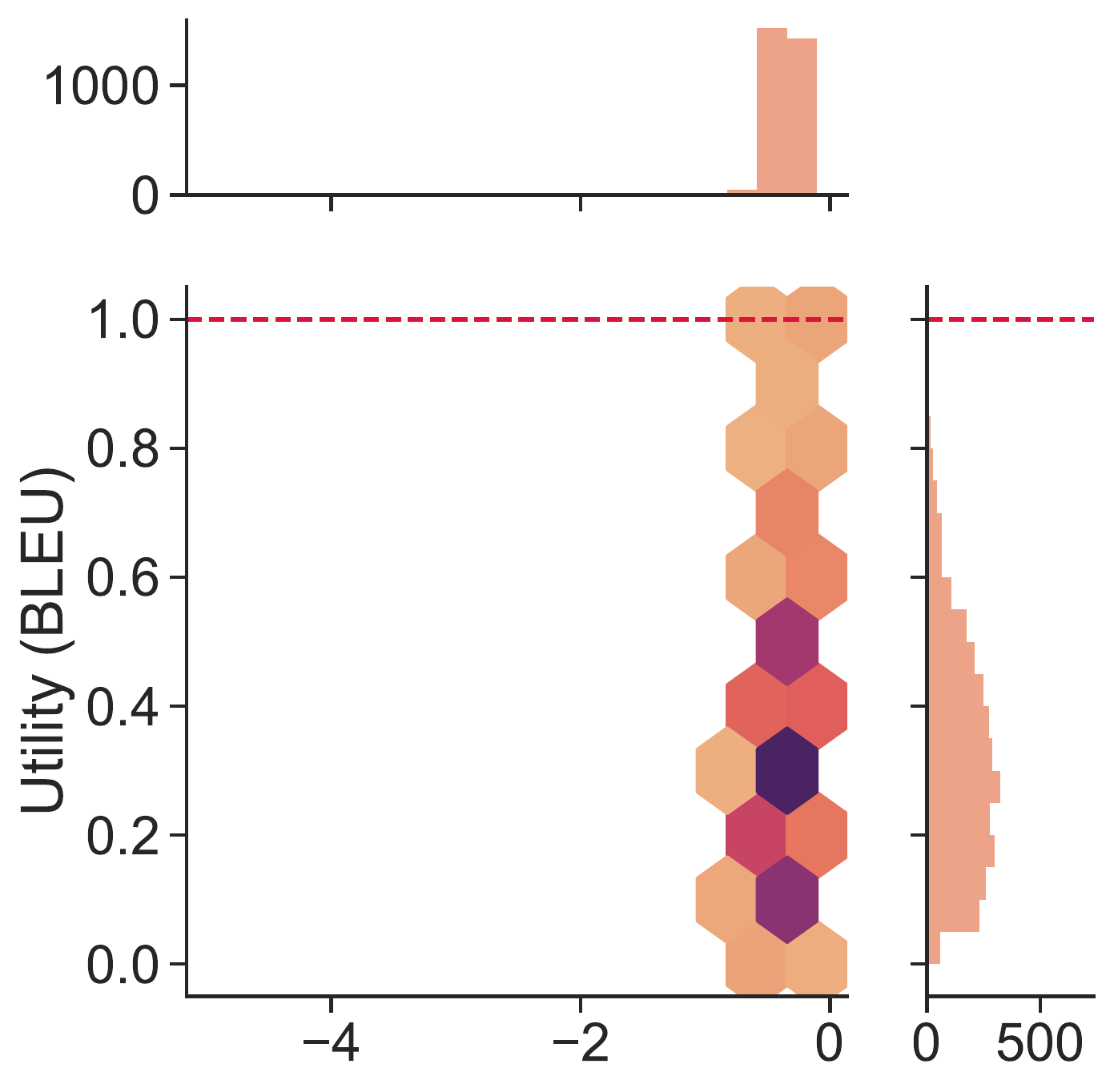}
         \label{fig:wmt14_bs_no_target_normalization}
     \end{subfigure}
     \begin{subfigure}[c]{0.2375\textwidth}
         \centering
         \includegraphics[width=.88\textwidth]{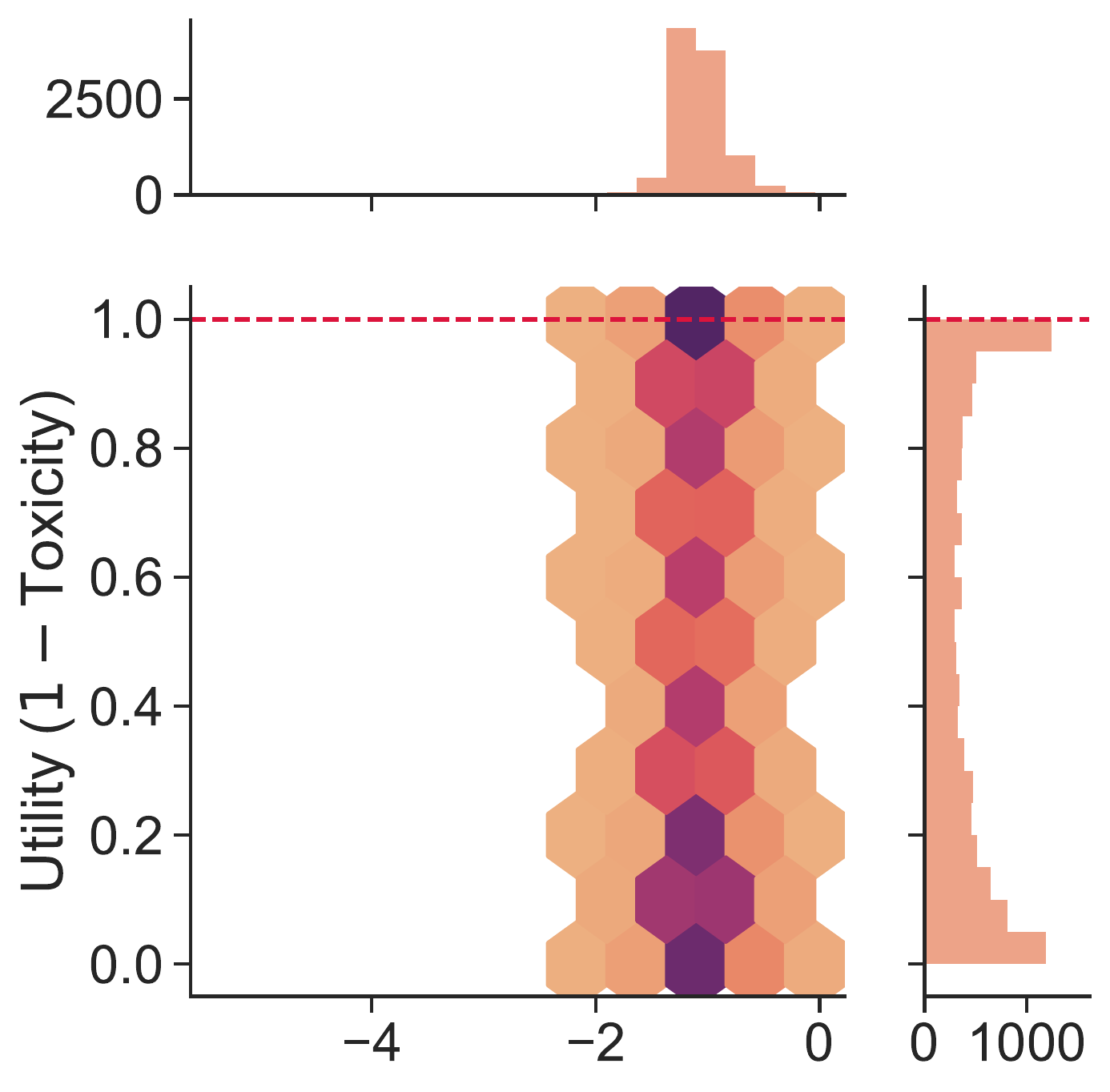}
         \label{fig:rtp_bs_no_target_normalization}
     \end{subfigure}
     \begin{subfigure}[c]{0.2375\textwidth}
         \centering
         \includegraphics[width=.88\textwidth]{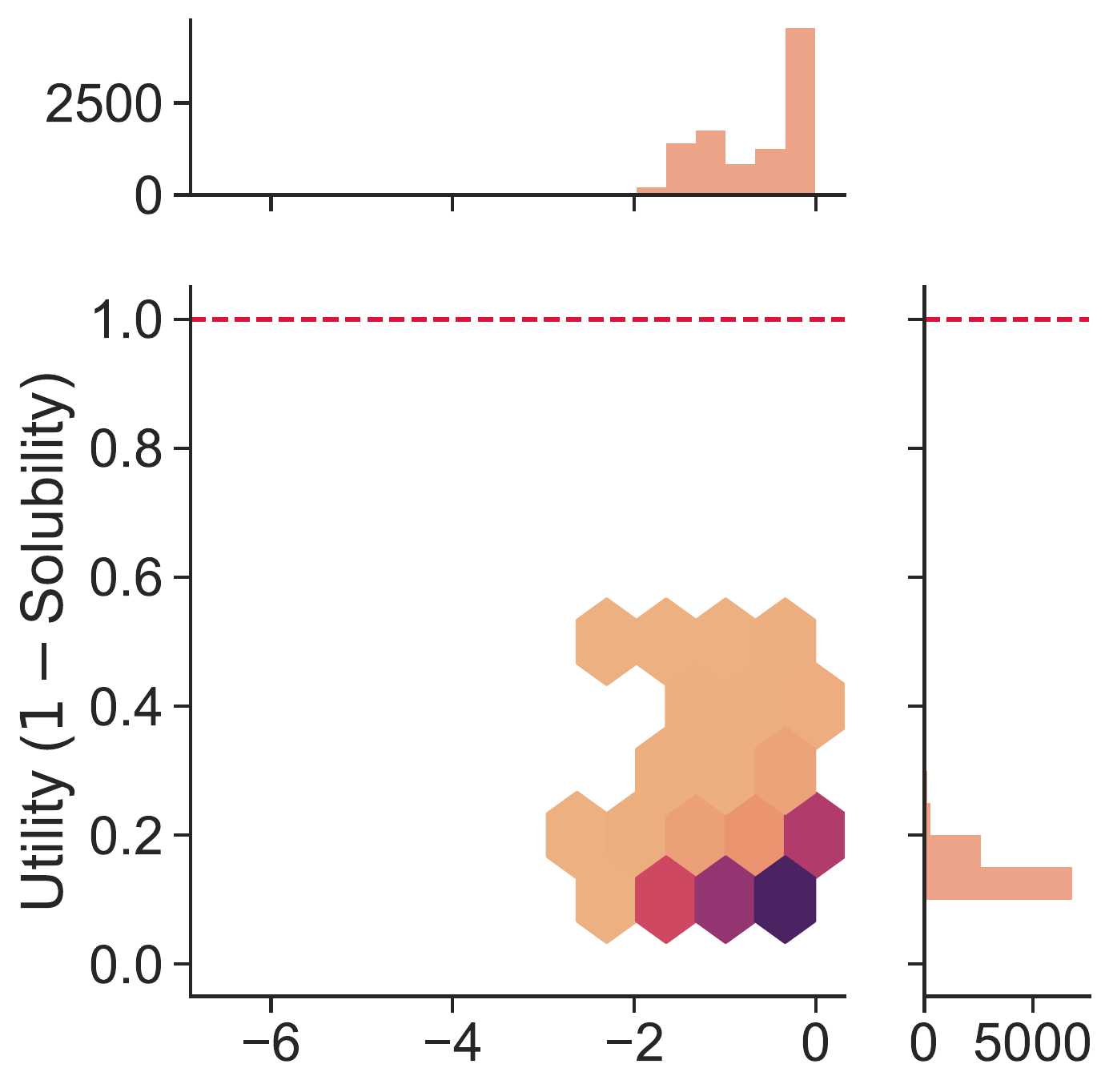}
         \label{fig:prot_bs_no_target_normalization}
     \end{subfigure}
     \rotatebox[origin=c]{90}{\footnotesize Stochastic Beams}
     \begin{subfigure}[c]{0.2375\textwidth}
         \centering
         \includegraphics[width=.88\textwidth]{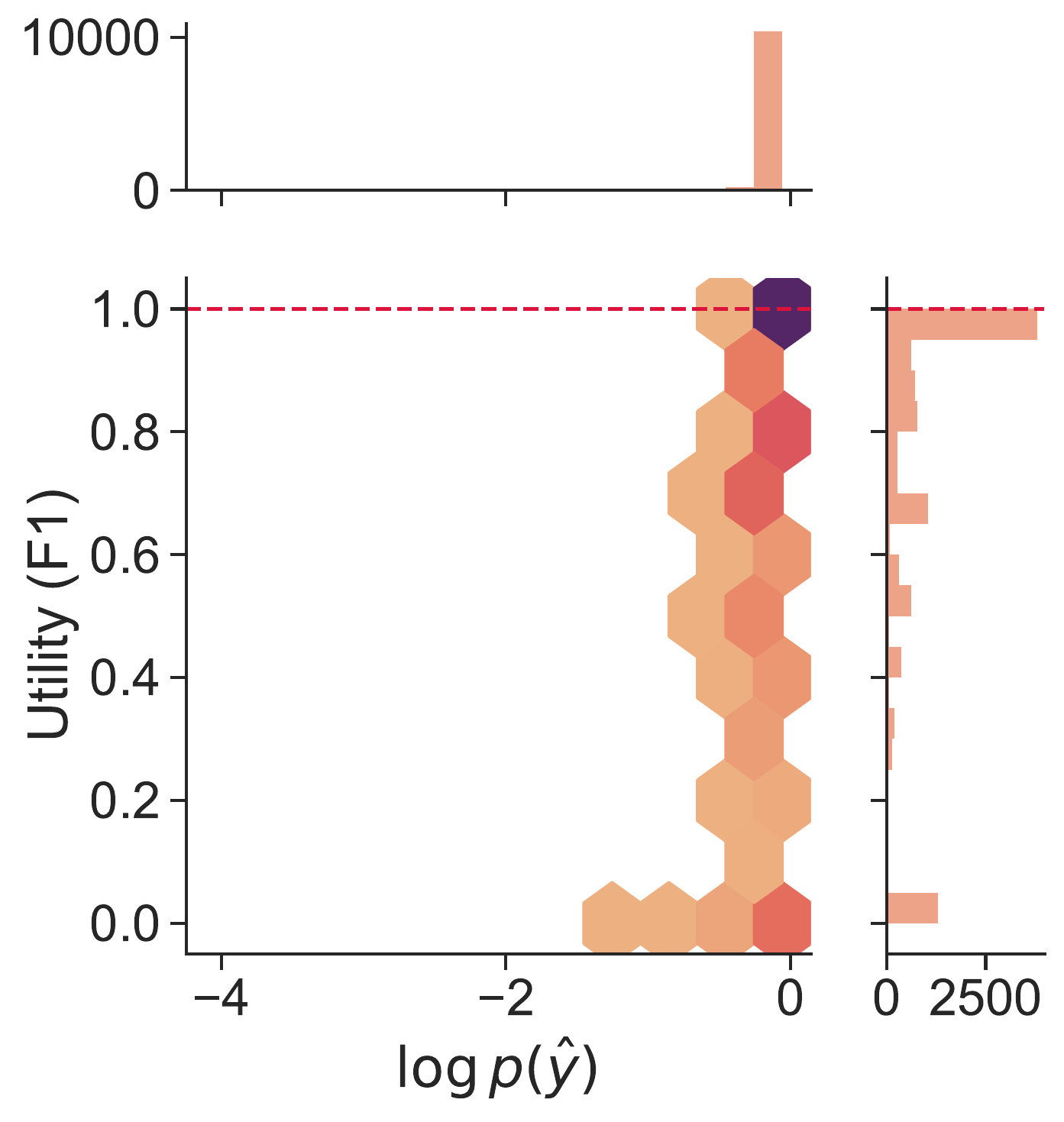}
         \caption{cIE (REBEL)}
         \label{fig:rebel_sbs_no_target_normalization}
     \end{subfigure}
     \begin{subfigure}[c]{0.2375\textwidth}
         \centering
         \includegraphics[width=.88\textwidth]{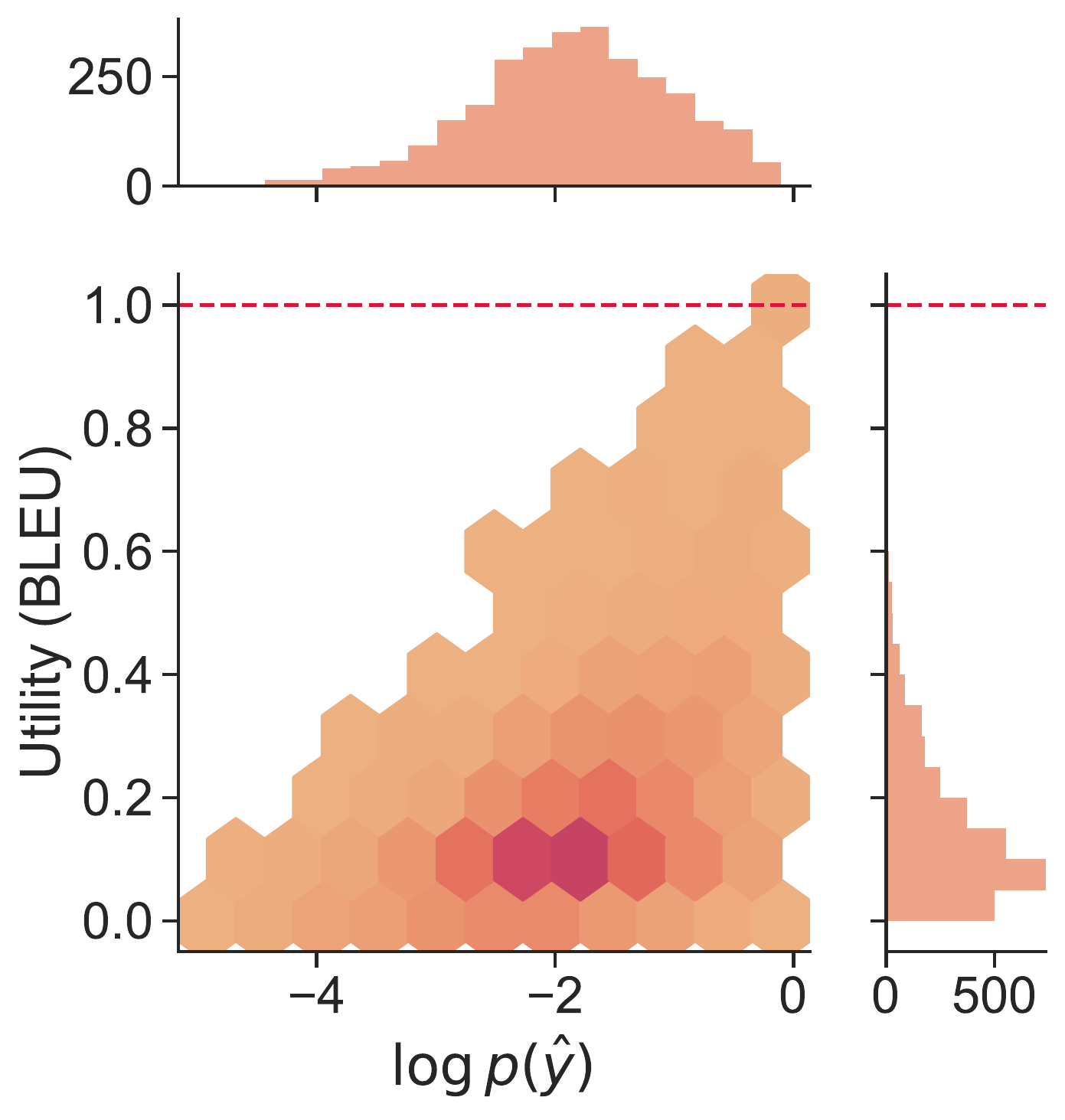}
         \caption{MT (WMT14)}
         \label{fig:wmt14_sbs_no_target_normalization}
     \end{subfigure}
     \begin{subfigure}[c]{0.2375\textwidth}
         \centering
         \includegraphics[width=.88\textwidth]{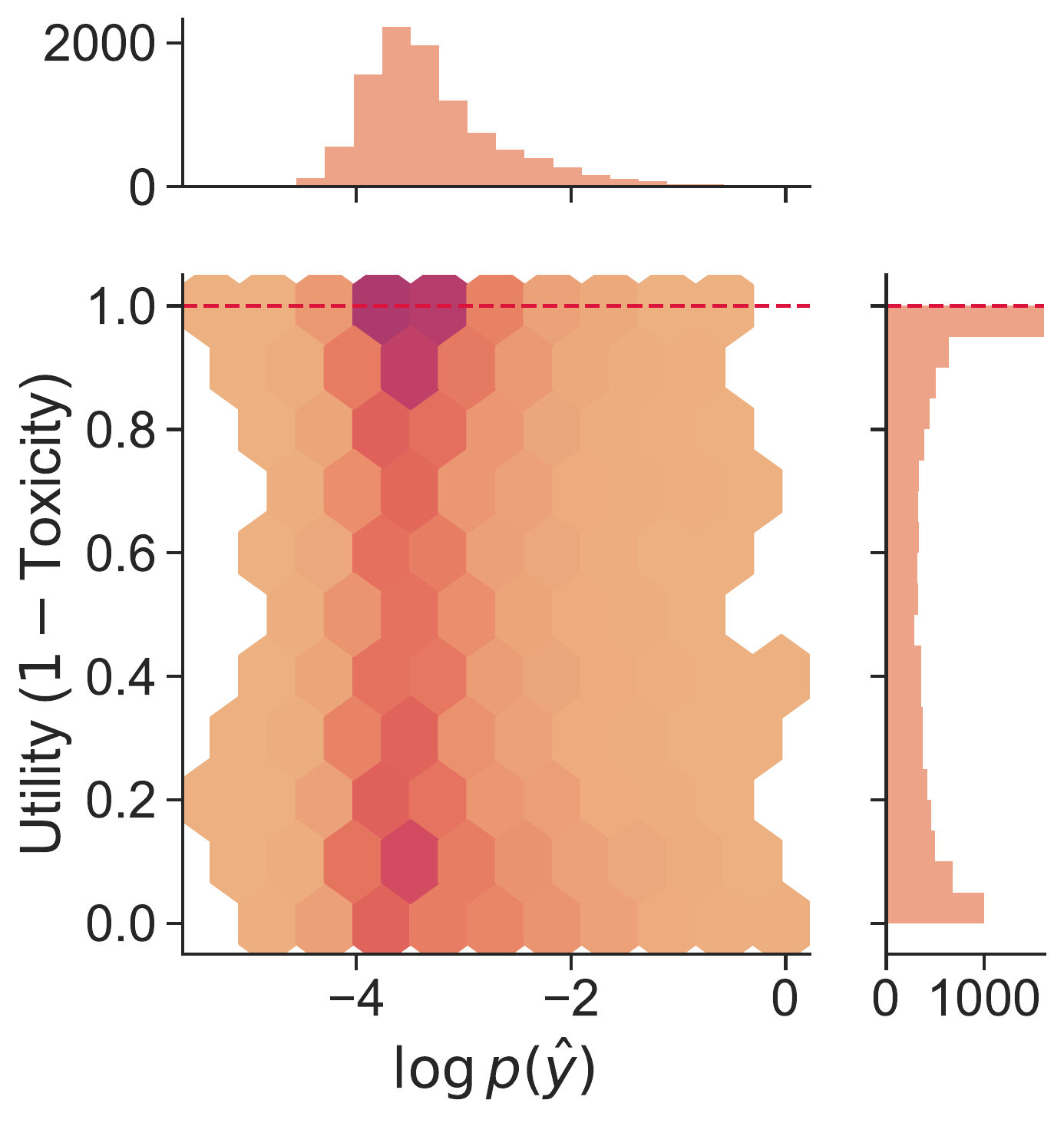}
         \caption{NTTG (RTP)}
         \label{fig:rtp_sbs_no_target_normalization}
     \end{subfigure}
     \begin{subfigure}[c]{0.2375\textwidth}
         \centering
         \includegraphics[width=.88\textwidth]{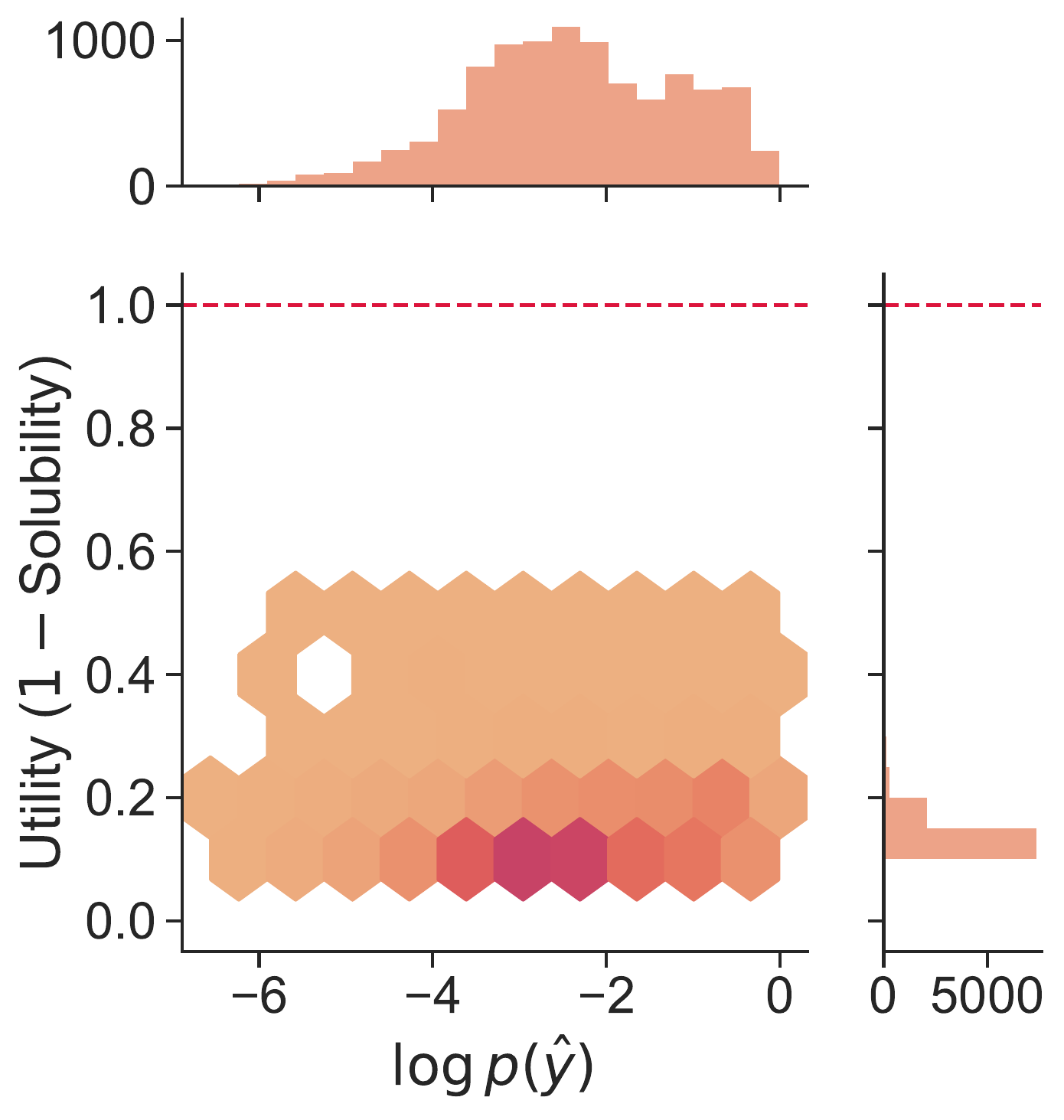}
         \caption{NSPG (SwissProt)}
         \label{fig:prot_sbs_no_target_normalization}
     \end{subfigure}
     \caption{\textbf{RQ1 (post-decoding alignment) -- without target normalization}: A version of \Figref{fig:fig1} without the target answer log-likelihood normalization.}
     \label{fig:fig1_no_target_normalization}
\end{figure*} 

\begin{table*}[!ht]
\centering
\resizebox{\linewidth}{!}{%
\begin{tabular}{@{}c|c|c@{}}
\toprule
      \textbf{ID} & \textbf{Input} & \textbf{Prediction}       \\ \midrule
1    & Is the following sentence plausible? Leon Draisaitl grounded out to second base in the National League Championship Series. A: & i'm pretty sure that's not really a grammatical sentence, although one might...
                        \\
2      & Is the following sentence plausible? John Collins threw a touchdown in the NFC divisional round. A: & the nor gates give you (not(a)) and (not(b))    \\
3 & Is the following sentence plausible? Jack Eichel dunked the ball. A: & hmm? well, i'm not really sure. let me look it up in \\\bottomrule
\end{tabular}%
}
\caption{Examples of outputs that are not providing an answer. The first and the second row provide an example where the model produces unrelated text, while the third row is an example of an indefinite answer.}
\label{tab:unparseable_examples}
\end{table*}

\subsection{Complexity Analysis of Decoding Algorithms} 
\label{app:complexity}
Most of the compute during decoding is allocated on querying the LM or the value model. Therefore, to show how decoding strategies compare in terms of the computation cost, in Table~\ref{tab:complexities} we provide a coarse complexity analysis in terms of the LM and value model calls.

\section{Experiments and Results}

\subsection{RQ1: The Likelihood–Utility Relationship}

\Figref{fig:fig1_no_target_normalization} is an alternative version \Figref{fig:fig1} where the x-axis (for cIE and MT) is not normalized using the log-likelihood of the target answers.

\subsection{RQ3: Prompting as an MMS}
\label{app:exp_and_results_rq3}

\subsubsection{Extracting Labels from Zero-Shot Predictions}
The outputs produced with zero-shot prompting do not follow a particular structure that can be used to extract the answer and therefore need to be processed manually. In some cases, it was not possible for an answer to be extracted. The two most common reasons for this were unrelated text as an answer or an indefinite answer. We provide examples of such predictions in \Tabref{tab:unparseable_examples}. Overall, 24.9\% of the answers could not be parsed. In such cases, we favored putting an indefinite label instead of "yes" or "no", and counting the answer as wrong irrespective of the ground truth label. If the answer and explanation were unrelated, but an answer was given, we did consider the answer.

\subsection{Computational Infrastructure and Runtime}
\label{app:compute_and_runtime}
The evaluation for RQ1 as well as the hyperparameter search for RQ2 were conducted on a single machine with 24 Intel(R) Xeon(R) CPU E5-2690 v4 @ 2.60GHz processor cores and 441 GB of RAM, equipped with 4 NVIDIA V100-PCIE-16GB GPUs. \Tabref{tab:running_time} provides the details for RQ1.

For VGBS, we performed a small hyperparameter search over different values for $\alpha={0.01, 0.25, 0.5, 0.75, 0.99}$ on 80 datapoints for each noise value of the value model. The same procedure was conducted for MCTS, over $c_{puct}={0.25, 1.25, 3}$. Each of these runs took 20 to 30 minutes of wall time, that is, slightly over 1 to 2 hours of GPU time. \Tabref{tab:vgbs_running_time_parameters} and \Tabref{tab:mcts_running_time_parameters} provide the final parameters for VGBS and MCTS respectively. 

The evaluation for RQ2 and RQ3, as well as the hyperparameter search for RQ2, were conducted on a single machine with 96 processor cores and 840 GB of RAM, equipped with 8 NVIDIA A100-SXM4-80GB GPUs. \Tabref{tab:vgbs_running_time_parameters} and \Tabref{tab:mcts_running_time_parameters} provide the details for RQ2.

The evaluation for RQ3 was performed following the Sports Understanding setup in \citet{DBLP:journals/corr/abs-2201-11903}, by taking the same in-context examples. We used greedy decoding for all of the prompting methods. The running time for all prompting experiments (ZS, FS, and CoT) was 6 GPU hours.

\begin{table}
\centering
\resizebox{0.8\columnwidth}{!}{
\setlength{\tabcolsep}{5pt}
\begin{tabular}{lcccc}
\toprule
{} &  Number of Beams & Time (in GPU hours) \\
\midrule
cIE + Greedy &      1 &  1.5 \\
cIE + BS     &     10 & 10.5 \\
cIE + SB     &     10 & 10.5 \\
\midrule
MT + Greedy &      1 & 0.5 \\
MT + BS     &      5 & 1.0 \\
MT + SB     &      5 & 1.5 \\
\midrule
NTTG + Greedy &      1 & 2.0 \\
NTTG + BS     &      5 & 3.5 \\
NTTG + SB     &      5 & 4.5 \\
\midrule
NSPG + Greedy &      1 & 3.5 \\
NSPG + BS     &      5 & 5.5 \\
NSPG + SB     &      5 & 7.0 \\
\bottomrule
\end{tabular}
}
\caption{\textbf{Parameters for the greedy likelihood-based decoding algorithms.} The default parameters for each model were used, and no hyperparameter search was conducted.}
\label{tab:running_time}
\end{table}

\begin{table}
\centering
\resizebox{0.8\columnwidth}{!}{
\setlength{\tabcolsep}{5pt}
\begin{tabular}{lcc}
\toprule
{} &  $\alpha$ &  Time (in GPU hours) \\
\midrule
MT ($\lambda=0.99$) &      0.01 &      4  \\
MT ($\lambda=0.5$) &      0.01 & 4  \\
MT ($\lambda=0.35$) &      0.25 &    4  \\
MT ($\lambda=0.25$) &           0.25 & 4  \\
MT ($\lambda=0.15$) &      0.25 & 4 \\ 
MT ($\lambda=0.01$) &   0.75 &  4  \\
\midrule
NTTG (oracle) &      0.25 & 32  \\
NTTG (1200 steps)     &      0.25 &  30 \\
NTTG (400 steps)     &      0.25 & 30 \\
NTTG (200 steps)     &      0.25 & 29  \\
\bottomrule
\end{tabular}
}
\caption{\textbf{Parameters for VGBS.} For all the experiments, the value models consider the top-$10$ tokens according to the likelihood. The BLEU to the true target is weighted by $\lambda$ (i.e., high $\lambda$ translates to high-quality value model).}
\label{tab:vgbs_running_time_parameters}
\end{table}

\begin{table}
\centering
\resizebox{0.8\columnwidth}{!}{
\setlength{\tabcolsep}{5pt}
\begin{tabular}{lcc}
\toprule
{} &  $c_{puct}$ &  Time (in GPU hours) \\
\midrule
MT ($\lambda=0.99$) &    1.25 & 41  \\
MT ($\lambda=0.5$) &    0.5 & 41 \\
MT ($\lambda=0.35$) &    0.5 & 40.5 \\
MT ($\lambda=0.25$) &    0.25 & 40  \\
MT ($\lambda=0.15$) &   0.25 & 40 \\ 
MT ($\lambda=0.01$) &   1.25 & 40  \\
\midrule
NTTG (oracle) &      1 &  125 \\
NTTG (1200 steps)     &      1 &      96  \\
NTTG (400 steps)     &      1 &           100  \\
NTTG (200 steps)     &      1 &           98  \\
\bottomrule
\end{tabular}
}
\caption{\textbf{Parameters for MCTS.} For all the experiments, at each node, we consider the top-$20$ tokens according to the likelihood and perform 50 simulations. The BLEU to the true target is weighted by $\lambda$ (i.e., high $\lambda$ translates to high-quality value model).}
\label{tab:mcts_running_time_parameters}
\end{table}

\end{document}